%% file: main.tex
\documentclass{article}

\ifdefined\pdfminorversion
\fi

    \PassOptionsToPackage{numbers, compress, sort}{natbib}
    \PassOptionsToPackage{table, svgnames, dvipsnames}{xcolor}

 \usepackage[main, final]{neurips_2026}

\newcommand{\mytitle}{Scaling Reasoning Efficiently via \\ Relaxed On-Policy Distillation}
\newcommand{\alg}{\textsc{Reopold}\xspace}

\usepackage{microtype}
\usepackage{graphicx}
\usepackage{algorithm}
\usepackage{algorithmic}
\usepackage{subcaption}
\usepackage{booktabs} 

\usepackage{hyperref}

\usepackage[utf8]{inputenc} 
\usepackage[T1]{fontenc}    
\usepackage{url}            
\usepackage{amsfonts}       
\usepackage{nicefrac}       
\usepackage{xcolor}         

\usepackage{amsmath}
\usepackage{amssymb}
\usepackage{mathtools}
\usepackage{amsthm}

\usepackage{arydshln}
\usepackage{enumitem}
\usepackage{lipsum}
\usepackage{wrapfig, makecell}

\usepackage{bm}
\usepackage{tcolorbox}
\usepackage{xspace}
\usepackage{caption}

\definecolor{darkpastelgreen}{rgb}{0.01, 0.75, 0.24}
\definecolor{electriccrimson}{rgb}{1.0, 0.0, 0.25}

\usepackage{pifont}
\newcommand{\cmark}{\ding{51}}

\usepackage[capitalize,noabbrev]{cleveref}

\theoremstyle{plain}
\newtheorem{theorem}{Theorem}[section]

\theoremstyle{definition}

\theoremstyle{remark}
\newtheorem{remark}[theorem]{Remark}

\hypersetup{
    pdftitle={Relaxed On-Policy Distillation},
    colorlinks=true,
    linkcolor=red,
    citecolor=RoyalBlue,
    filecolor=cyan,
    urlcolor=Magenta
}
\title{\mytitle}

\author{%
    Jongwoo Ko \quad Sara Abdali \quad Young Jin Kim \quad Tianyi Chen \quad Pashmina Cameron \\
    \url{https://relaxed-opd.github.io}
}

\begin{document}

\maketitle

\begin{abstract}
On-policy distillation (OPD) is a promising approach for transferring reasoning capabilities to capacity-constrained models, yet sampled-token OPD often suffers from entropy collapse and negative transfer. We identify uniform credit allocation as a key bottleneck: existing objectives obtain dense teacher feedback on student-generated trajectories, but apply it uniformly across sampled tokens despite highly heterogeneous token-level signals. Most tokens carry redundant near-zero credit, while rare heavy-tailed negative credits can dominate updates and prematurely suppress plausible reasoning trajectories. We introduce \alg (\textbf{Re}laxed \textbf{O}n-\textbf{Pol}icy \textbf{D}istillation), a framework that relaxes this uniform assignment by controlling where teacher feedback is applied, how strongly it affects the update, and when the allocation rule changes during training. Across diverse reasoning tasks, \alg improves sampled-token OPD over recent post-training baselines with up to 12$\times$ higher sample efficiency, and extends to cross-vocabulary distillation, self-distillation, and test-time scaling.
\end{abstract}

\input{sections/introduction}
\input{sections/motivation}
\input{sections/approach}
\input{sections/experiments}
\input{sections/conclusion}

\bibliographystyle{plainnat}
\bibliography{main}


\clearpage
\appendix
\input{sections/related-work-extension}
\input{sections/proof}
\input{sections/algorithm}
\input{sections/experimental-setup}

\input{sections/extended_experiments}
\input{sections/qualitative}



\end{document}

%% file: sections/introduction.tex
\vspace{-10pt}
\section{Introduction}
Large language models (LLMs) have achieved remarkable reasoning capabilities through large-scale post-training and test-time scaling, as demonstrated by recent reasoning-oriented systems such as OpenAI’s o1/o3 \citep{jaech2024openai} and DeepSeek-R1 \citep{guo2025deepseek}. However, replicating this success in small language models (SLMs) proves difficult. Due to limited capacity, SLMs often benefit less efficiently from direct reward optimization, making standard reinforcement learning (RL) less effective for improving reasoning in capacity-constrained models~\cite{guo2025deepseek, dang2025reinforcement, yan2025learning}. This disparity necessitates alternative mechanisms specifically tailored for transferring reasoning abilities to capacity-constrained models.

To address this, recent work~\citep{yang2025qwen3, patiño2025unlocking} has adopted on-policy distillation (OPD), where the student learns from its own trajectories under the guidance of a high-capacity teacher with stronger reasoning capabilities. By providing dense teacher-derived feedback on student-sampled trajectories, OPD reduces the burden of reward design while preserving the on-policy nature of training. Empirically, OPD has been shown to outperform RL and supervised fine-tuning (SFT) on reasoning tasks, making it a promising framework for transferring reasoning capabilities. More recently, sampled-token supervision~\citep{lu2025onpolicy, xiao2026mimo} has emerged as a common alternative to full-vocabulary distillation~\citep{agarwal2024onpolicy, ko2024distillm}, as it avoids scoring the entire vocabulary and therefore scales more naturally to long reasoning trajectories.

However, a central bottleneck in OPD lies not only in acquiring teacher feedback, but also in allocating it effectively. Existing sampled-token methods compute teacher-derived credits (\textit{i.e.,} log-likelihood ratios) on student-generated trajectories, but largely apply these credits uniformly over all sampled tokens. This uniform allocation overlooks the highly imbalanced nature of token-level credits: most sampled tokens provide redundant near-zero signals, whereas rare but heavy-tailed negative credits can dominate optimization, as shown in \autoref{fig:rewards}. As a result, stable and efficient OPD requires mechanisms that decide \textit{where} teacher feedback should be applied, \textit{how much} it should influence the update, and \textit{when} it should be trusted.

Empirically, this imbalance manifests as instability in standard OPD~\citep{gudibande2023false, gu2024minillm}. We observe frequent negative transfer, where the student degrades relative to its base initialization (Section~\ref{sec:math} and \autoref{fig:math-7b}), along with rapid entropy collapse that leads to premature convergence (Section~\ref{sec:motivation} and \autoref{fig:multi}). This motivates our selective credit allocation view, which is related to prior work on non-uniform sample allocation and robust example reweighting~\citep{schaul2015prioritized, shrivastava2016training, lin2017focal, ren2018learning}. Rather than treating all sampled-token credits uniformly, OPD should selectively allocate teacher supervision according to the student's current learning dynamics.

\input{figures/figure_intro}
\vspace{-5pt}
\paragraph{Contributions.} We analyze sampled-token OPD as a problem of selective credit allocation. Instead of treating all teacher-derived token signals as equally useful, we ask where teacher feedback should be applied, how strongly it should influence the update, and when the allocation rule should change during training. Our contributions are summarized as follows:
\begin{itemize}[leftmargin=*, itemsep=0pt]
    \vspace{-5pt}
    \item \textbf{Credit allocation diagnosis:} We show that standard OPD implicitly applies a uniform credit allocation rule over sampled tokens. This rule is poorly matched to the heterogeneous credit structure of OPD, where a large fraction of sampled tokens carry near-zero credit, while rare teacher-student mismatches induce heavy-tailed negative credit and premature entropy collapse.
    \item \textbf{Student-conditioned credit allocation:} To address these, we propose \textbf{\alg}, a unified framework that selectively regulates teacher credit along three axes: update magnitude, token uncertainty, and training stage. Concretely, \alg integrates bounded negative credit, entropy-guided token selection, and exploration-to-refinement scheduling to stabilize sampled-token distillation.
    \item \textbf{State-of-the-art efficiency and scalability:} Empirically, \alg achieves superior training sample efficiency and enables stronger test-time scaling for SLMs (\autoref{fig:teaser}). Across mathematical, visual \cite{liu2025noisyrollout}, and tool-use \cite{qian2025toolrl} reasoning tasks, \alg consistently outperforms baselines and remains effective beyond same-family distillation, including cross-vocabulary \cite{patiño2025unlocking, boizard2025towards} and privileged self-distillation \cite{zhao2026self}.
\end{itemize}

%% file: figures/figure_intro.tex
\begin{figure}[t]
    \centering
    \includegraphics[width=1.0\linewidth]{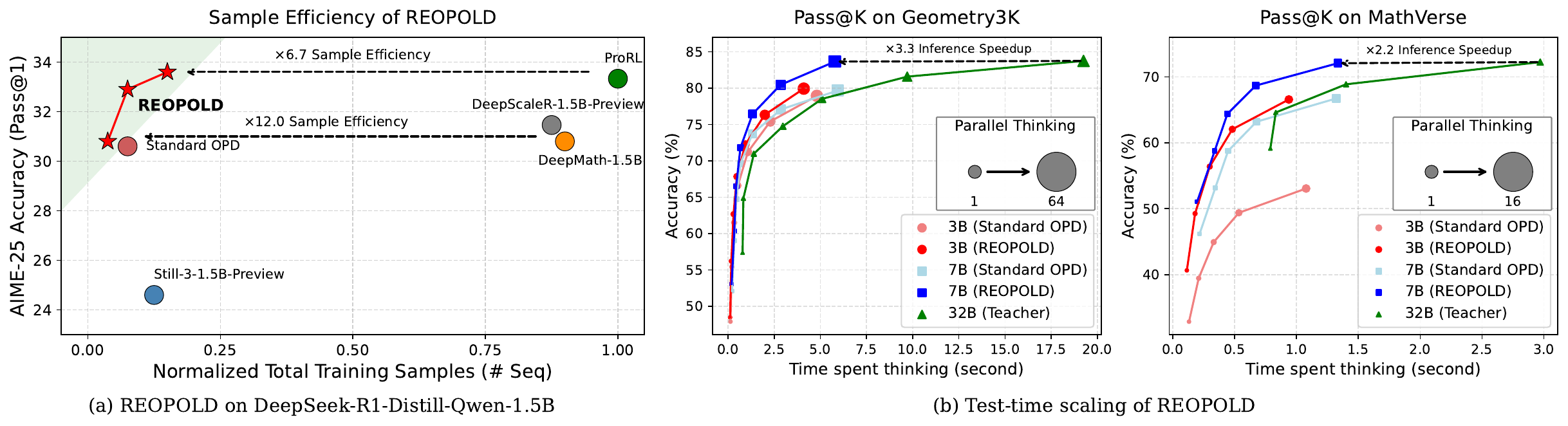}
    \caption{
        \textbf{Performance of \alg.} (a) \textbf{Sample Efficiency}: \alg achieves a favorable accuracy–sample-efficiency trade-off on AIME-25 compared with existing post-training baselines, where sample efficiency is measured by the number of generated training sequences. Detailed explanation can be found in Section~\ref{sec:math}. (b) \textbf{Test-time Scaling}: On visual reasoning tasks, \alg demonstrates superior test-time scaling capabilities compared to the standard OPD baseline. Notably, it allows smaller models to approach the performance of the 32B teacher. Detailed explanation can be found in Section~\ref{sec:visual}.
    }
    \vspace{-12.5pt}
    \label{fig:teaser}
\end{figure}

%% file: sections/motivation.tex
\vspace{-10pt}
\section{Motivation: Indiscriminate Credit Allocation in On-Policy Distillation}\label{sec:motivation}

\vspace{-5pt}
\subsection{Background: Sampled-token on-policy distillation}
Let $\pi_\theta$ denote the student policy and $\pi_T$ the teacher policy. For a query $q$, OPD samples responses from the student's behavior policy $\pi_{\theta_{\mathrm{old}}}$ and updates the student using teacher feedback on the sampled tokens. A common sampled-token reverse-KL objective, written in maximization form, is
\begin{equation}
\label{eq:rkl_off}
\begin{split}
    \mathcal{J}_{\mathrm{RKL}}(\theta)
    =
    \mathbb{E}_{q \sim \mathcal{Q},\, \{o_i\}_{i=1}^{G} \sim \pi_{\theta_{\mathrm{old}}}(\cdot|q)}
    \left[
    \frac{1}{\sum_i |o_i|}
    \sum_{i=1}^{G}\sum_{t=1}^{|o_i|}
    \rho_{i,t}(\theta) R_{i,t}(\theta)
    \right],
\end{split}
\end{equation}
where
\begin{equation}\label{eq:credit}
    \rho_{i,t}(\theta)
    =
    \frac{\pi_\theta(o_{i,t}|q,o_{i,<t})}
    {\pi_{\theta_{\mathrm{old}}}(o_{i,t}|q,o_{i,<t})},
    \qquad
    R_{i,t}(\theta)
    =
    \log
    \frac{\pi_T(o_{i,t}|q,o_{i,<t})}
    {\pi_\theta(o_{i,t}|q,o_{i,<t})}.
\end{equation}
We refer to $R_{i,t}$ as the teacher-derived token credit. Following sampled-token OPD practice, we detach $R_{i,t}$ and treat it as a fixed token-level credit within each update; the corresponding gradient derivation is provided in Appendix~\ref{app:proof}.

This sampled-token formulation differs from full-vocabulary distillation~\citep{agarwal2024onpolicy, gu2024minillm, ko2024distillm}, which compares teacher and student distributions over the entire vocabulary at each generated position. While full-vocabulary supervision provides richer feedback, it is memory-intensive for modern reasoning distillation with long trajectories and multiple rollouts (\autoref{tab:full-vocab}). A lightweight alternative is to evaluate the teacher only on the token sampled by the student, which has become widely used in recent OPD work~\citep{lu2025onpolicy, xiao2026mimo, yang2026learning}. We also provide detailed related work in Appendix~\ref{app:related} and empirical comparisons with full-vocabulary OPD and their corresponding Top-$K$ approximations in Table~\ref{tab:full-vocab}.

\input{figures/rewards2}

\vspace{-5pt}
\subsection{Diagnosing Credit Misallocation in Sampled-Token OPD}
\label{sec:credit_allocation_theory}
Sampled-token OPD provides teacher feedback only on tokens sampled by the student, but standard objectives (Eq. \ref{eq:rkl_off}) aggregate these credits uniformly across samples. This treats all sampled tokens as equally useful supervision targets, despite many credits being uninformative, redundant, or extreme. Because token-level credits are highly skewed in practice, this uniform allocation creates two finite-sample failure modes.

\vspace{-7.5pt}
\paragraph{Most token credits are redundant.}
As shown in \autoref{fig:rewards}, the sampled-token credit distribution has a large mass near zero. These credits occur when the teacher and student assign similar probability to the sampled token, yielding little corrective signal. While such tokens are not harmful individually, uniform aggregation makes them consume normalization mass and mini-batch budget that could be allocated to more informative tokens. As a result, many updates are diluted by nearly inactive credits, reducing the effective learning signal obtained from each sampled batch.

\vspace{-7.5pt}
\paragraph{A few negative credits dominate the update.} 
The more severe failure mode arises from the negative tail of the credit distribution. As shown in \autoref{fig:rewards}, when the student samples a token that the teacher assigns very low probability, the credit $R_{i,t}(\theta)$ in Eq.~\ref{eq:credit} can become highly negative. Although this tail accounts for only a small fraction of sampled positions, its large magnitude can dominate optimization. To measure this concentration, we compute a logit-level update-energy proxy,
\begin{equation}\label{eq:energy}
    E_{i,t}(\theta)
    =
    R_{i,t}(\theta)^2
    \left(
    1-\|\pi_\theta(\cdot \mid q,o_{i,<t})\|_2^2
    \right),
\end{equation}
which serves as an expected logit-level proxy for the squared gradient and allows us to decompose update energy by credit quantile. This proxy assumes stop-gradient through $R_{i,t}$, replacing the token-specific score norm by its expectation $1-\|\pi_\theta\|_2^2$; see Appendix~\ref{app:energy} for the derivation. We find that the bottom $1\%$ most negative credits account for over $95\%$ of the total update energy in early standard OPD training (\textcolor{red}{Table~}\ref{tab:energy}).

This concentration can be harmful because large negative credits strongly suppress the probability of sampled tokens \cite{wu-etal-2025-rethinking, ko2025distillm}. In autoregressive models, suppressing an intermediate token can also disrupt the subsequent reasoning path conditioned on it. Thus, standard OPD may reduce support for trajectories that are locally disfavored by the teacher but still useful for exploration. We therefore use a base-policy retention diagnostic to measure how much support later checkpoints assign to trajectories sampled from the initial student $\pi_{\theta_0}$:
\begin{equation}\label{eq:retention}
    \mathrm{Ret}(k)=
    \mathbb{E}_{q,o\sim\pi_{\theta_0}}
    \left[
    \frac{1}{|o|}\sum_t \log \pi_{\theta_k}(o_t|q,o_{<t})
    \right].
\end{equation}
Lower values mean that the current policy assigns less probability to trajectories originally generated by the base policy. Standard OPD rapidly reduces this support: within the first 10 steps, retention decreases from -1.58 and -1.43 to -2.55 and -2.32 (\textcolor{red}{Table~}\ref{tab:retention}), corresponding to about 40\% of the initial token-level likelihood. Together with the update-energy concentration and entropy collapse in \autoref{fig:multi}, this suggests rapid policy concentration under standard OPD.

\vspace{-7.5pt}
\paragraph{Useful teacher signal is student-conditioned.}
These observations show that teacher feedback is not uniformly useful across sampled tokens. Near-zero credits provide little corrective signal, while rare large negative credits may capture real teacher-student disagreement but destabilize optimization at full strength. The bottleneck in standard OPD is therefore credit allocation: which sampled-token credits to use, how strongly to apply them, and when to change the allocation rule during training.

%% file: figures/rewards2.tex
\begin{figure}[t]
\centering
\begin{minipage}[t]{0.59\linewidth}
    \centering
    \small
    \includegraphics[width=\linewidth,height=0.42\textheight,keepaspectratio]{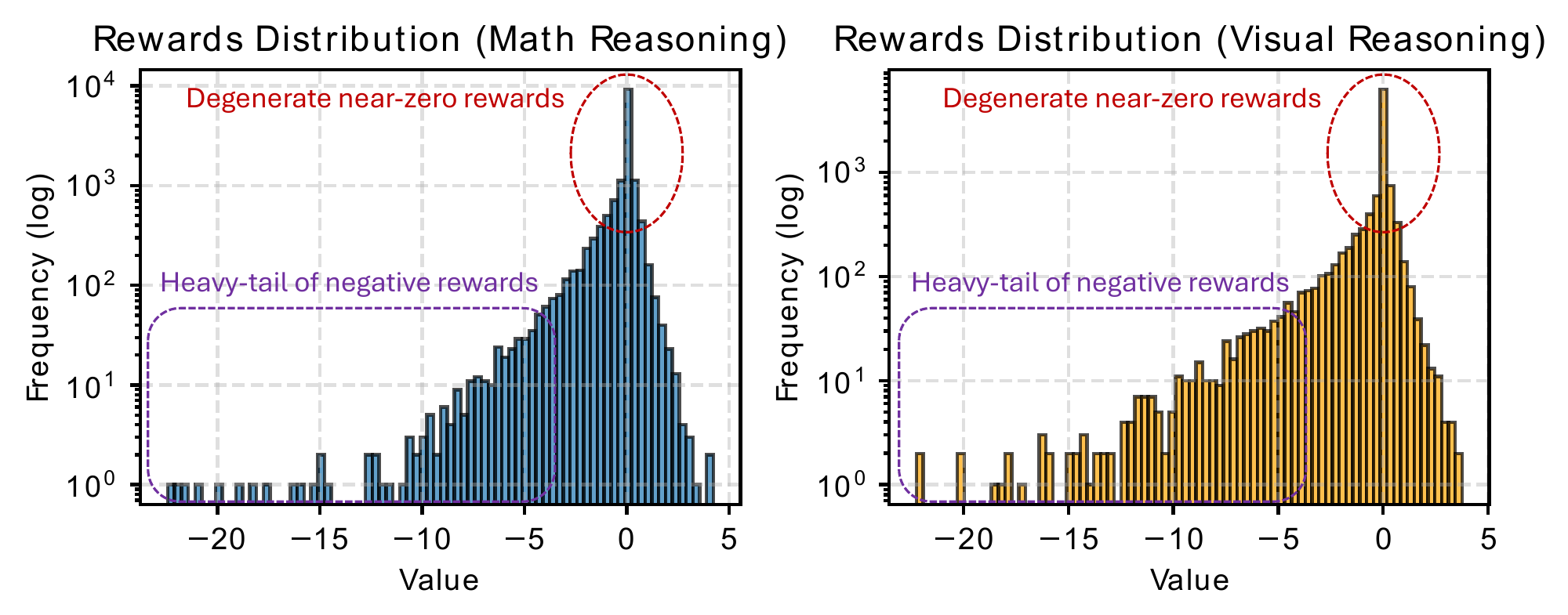}
    \caption{\textbf{Log-scale histogram of teacher-derived credit.} The \textcolor{red}{red dashed ellipse} indicates degenerate near-zero credits, while the \textcolor{purple!90}{purple dashed region} highlights the heavy tail of negative credits.}
    \label{fig:rewards}
\end{minipage}
\hfill
\begin{minipage}[t]{0.37\linewidth}
    \vspace{-87.5pt}
    \centering
    \small
    \captionof{table}{Update-energy (Eq.~\ref{eq:energy}) concentration in negative-credit tails.}
    \label{tab:energy}
    \vspace{-5pt}
    \resizebox{\linewidth}{!}{
    \begin{tabular}{lccc}
        \toprule
        Bottom & $\sim$1\% & $\sim$5\% & $\sim$10\% \\
        \midrule
        Math & 95.9\% & 97.1\% & 98.4\% \\
        Visual & 97.8\% & 98.6\% & 99.5\% \\
        \bottomrule
    \end{tabular}
    }
    \captionof{table}{Retention (Eq.~\ref{eq:retention}) on base policy samples during OPD training.}
    \label{tab:retention}
    \resizebox{\linewidth}{!}{
    \begin{tabular}{lccc}
        \toprule
        Step ($k$) & 0 & 10 & 300 \\
        \midrule
        Math & -1.583 & -2.554 & -2.841 \\
        Visual & -1.427 & -2.315 & -2.741 \\
        \bottomrule
    \end{tabular}
    }
\end{minipage}
\vspace{-10pt}
\end{figure}

%% file: sections/approach.tex
\vspace{-5pt}
\section{\alg: Relaxed On-Policy Distillation via Selective Credit Allocation}
The analysis in Section~\ref{sec:motivation} motivates a selective credit allocation view of OPD. Under this view, an objective should decide \textit{where} teacher feedback should be applied, \textit{how much} credit should be allowed to influence the update, and \textit{when} negative correction should be trusted. \alg instantiates these choices through \textcolor{darkpastelgreen}{entropy-guided token selection}, \textcolor{blue}{bounded negative credit}, and an \textcolor{electriccrimson}{exploration-to-refinement schedule}. For each batch $\mathcal{B}$ sampled from the query set $\mathcal{Q}$, we maximize:
\begin{equation}\label{eq:unified_objective}
    \mathcal{J}_{\text{\alg}}(\theta) = \mathbb{E}_{\mathcal{B} \sim \mathcal{Q}, q \sim \mathcal{B}, \, \{ o_i \}_{i=1}^{G} \sim \pi_{\theta_\mathrm{old}}(\cdot \mid q)} \left[ \frac{1}{\sum_{i=1}^{G}\textcolor{electriccrimson}{\sum_{t=1}^{|o_i|} M_{i,t}^{(k)}}} \sum_{i=1}^{G}\sum_{t=1}^{|o_i|} \rho_{i,t}(\theta) \textcolor{blue}{\hat{R}_{i,t}^{\lambda}(\theta)} \textcolor{electriccrimson}{M_{i,t}^{(k)}} \right],
\end{equation}
where
\begin{equation}
    \textcolor{blue}{\hat{R}_{i,t}^{\lambda}(\theta)} = \max \left( R_{i,t}(\theta), \frac{\log \lambda}{1-\lambda}\right),
    \qquad
    \textcolor{electriccrimson}{M_{i,t}^{(k)}} = 
    \begin{cases} 
        \mathbb{I}\left[ \textcolor{blue}{R_{i,t}(\theta)} \ge \frac{\log \lambda}{1-\lambda} \right] & \text{if } k < T_{\text{switch}} \\
        \mathbb{I}\left[ \textcolor{darkpastelgreen}{H_{t}^i} \ge \tau_\beta \right] & \text{if } k \ge T_{\text{switch}}
    \end{cases}
\end{equation}
Throughout training, we treat the teacher-derived credits and token-selection masks as stop-gradient quantities when computing policy updates. The full training procedure is summarized in Algorithm~\ref{alg:main}.

\vspace{-5pt}
\subsection{Where to Allocate Credit: Entropy-Guided Token Selection}\label{sec:token-level-dynamic-sampling}
\input{figures/entropy}
The first axis of credit allocation concerns \textit{where} teacher feedback should be applied. Standard OPD allocates update budget uniformly across all sampled tokens, even though many near-zero credit tokens occur at low-entropy positions. As shown in \autoref{fig:entropy}, these tokens are concentrated in contexts where the student and teacher assign similar probabilities to the sampled token, making additional correction largely redundant. The following elementary softmax identity supports using entropy as a lightweight allocation signal.

\begin{remark}\label{prop:softmax_score_entropy}
\textit{Let $z\in\mathbb{R}^{|\mathcal{V}|}$ be the logit vector over a finite vocabulary
$\mathcal{V}$, and let $p=\operatorname{softmax}(z)$ be the categorical distribution, with $p_v$ denoting the probability of token
$v\in\mathcal{V}$. If $y\sim p$, then}
\begin{equation}\label{eq:softmax_score_entropy}
    \mathbb{E}_{y\sim p}
    \left[
    \|\nabla_z \log p_y\|_2^2
    \right]
    =
    1-\|p\|_2^2
    \le H(p),
\end{equation}
\textit{where $p_y$ is the probability of the sampled token and $H(p)=-\sum_{v\in\mathcal{V}}p_v\log p_v$.}
\end{remark}

Applied to $p=\pi_\theta(\cdot|x)$ at context $x=(q,o_{<t})$, 
$\nabla_z \log p_y$ is the logit-level score of the sampled token, which captures how much a token-level likelihood update can locally redistribute probability mass~\citep{williams1992simple, kakade2001natural}. 
Remark~\ref{prop:softmax_score_entropy} shows that entropy upper-bounds the expected squared score norm, which captures local update leverage in logit space; the derivation is given in Appendix~\ref{app:softmax_score_identity}. 
The teacher credit provides the direction and magnitude of supervision, while student entropy serves as a proxy for local update leverage, motivating its use as a token-selection signal.

Motivated by this, we use student entropy as a lightweight allocation signal. For each sampled token $o_{i,t}$, let $H_t^i$ denote the entropy of the student's policy $\pi_\theta(\cdot|q,o_{i,<t})$. We define a token-level allocation mask
\begin{equation}
    M_{i,t}^{\mathrm{ent}}
    =
    \mathbb{I}\left[H_t^i \ge \tau_\beta\right],
\end{equation}
where $\tau_\beta$ is chosen so that $M_{i,t}^{\mathrm{ent}}$ selects the top-$\beta$ fraction of token entropies within the batch. This mask selects positions where the student remains uncertain, rather than uniformly accepting teacher credit at every sampled token. Integrating this mask into the sampled-token objective gives
\begin{equation}
\label{eq:entropy_mask}
\begin{split}
    \mathcal{J}_{\mathrm{Ent}}(\theta)
    =
    \mathbb{E}_{\mathcal{B} \sim \mathcal{Q}, q \sim \mathcal{B}, \{o_i\}_{i=1}^{G} \sim \pi_{\theta_{\mathrm{old}}}(\cdot|q)}
    \left[
    \frac{
    \sum_{i=1}^{G}\sum_{t=1}^{|o_i|}
    \rho_{i,t}(\theta)
    R_{i,t}(\theta)
    M_{i,t}^{\mathrm{ent}}
    }{
    \sum_{i=1}^{G}\sum_{t=1}^{|o_i|}
    M_{i,t}^{\mathrm{ent}}
    }
    \right].
\end{split}
\end{equation}
This creates a dynamic token batch whose effective size is determined by the student's uncertainty rather than sequence length. By filtering out low-entropy, near-zero-credit tokens, entropy-guided allocation reduces gradient dilution and concentrates update budget on positions where teacher feedback is most likely to be useful. Empirically, the sensitivity analysis in \autoref{fig:entropy_3b_7b} confirms that stricter entropy selection accelerates convergence and improves final accuracy, supporting the view that selective allocation of teacher credit is more effective than uniform token-level imitation.

\vspace{-5pt}
\subsection{How Much Credit to Use: Bounded Negative Credit}\label{sec:reward_clipping}
The second allocation axis controls how strongly the student should absorb negative teacher feedback. If the student samples a token with negligible teacher probability, the log-ratio credit can become extremely negative and dominate the update. We bound this tail using a mixture relaxation of the teacher and student distributions. For $\lambda\in[0,1)$, Jensen's inequality gives
\begin{equation}
\label{eq:mixture_bound_main}
\begin{split}
    R_{i,t}(\theta)
    =
    \log \frac{\pi_T(o_{i,t}|q,o_{i,<t})}
    {\pi_{\theta}(o_{i,t}|q,o_{i,<t})}
    \leq
    \frac{1}{1-\lambda}
    \log
    \frac{
    (1-\lambda)\pi_T(o_{i,t}|q,o_{i,<t})
    +
    \lambda\pi_{\theta}(o_{i,t}|q,o_{i,<t})
    }
    {\pi_{\theta}(o_{i,t}|q,o_{i,<t})}.
\end{split}
\end{equation}

The key insight comes from the asymptotic behavior of the relaxed term; detailed derivation is provided in Appendix~\ref{app:clip_proof}. While the original credit $R_{i,t}(\theta)$ diverges to $-\infty$ as $\pi_T(o_{i,t}|q,o_{i,<t})\rightarrow 0$, the relaxed expression converges to $c_\lambda = \frac{\log \lambda}{1-\lambda}$. Motivated by this asymptotic limit, we use $c_\lambda$ as a clipping floor for the original sampled-token credit, rather than directly optimizing the mixture-relaxed objective:
\begin{equation}
\label{eq:clip}
    \hat{R}_{i,t}^{\lambda}(\theta)
    =
    \max
    \left(
    R_{i,t}(\theta),
    c_\lambda
    \right).
\end{equation}
In the credit allocation view, this operation decides how much a single negative teacher signal is allowed to influence the student. It preserves the objective for ordinary tokens, while preventing rare teacher-student mismatches from consuming disproportionate update budget. Unlike Skew RKL \cite{ko2024distillm}, which changes the full objective and is sensitive to $\lambda$, this rule only targets outlying negative credits, yielding the robustness shown in \autoref{fig:clip_sensitivity}.

\subsection{When to Change Allocation: Exploration-to-Refinement Scheduling}\label{sec:multi_stage}

The third allocation axis controls \textit{when} different teacher credits should be accepted. The usefulness of teacher feedback changes over training: early updates should preserve plausible student trajectories, while later updates should refine unresolved token decisions. We implement this temporal control through a token-wise allocation mask:
\begin{equation}
\label{eq:masked_objective}
\begin{split}
    \mathcal{J}_{\text{\alg}}(\theta)
    =
    \mathbb{E}_{\mathcal{B} \sim \mathcal{Q}, q \sim \mathcal{B}, \{ o_i \}_{i=1}^{G} \sim \pi_{\theta_\mathrm{old}}(\cdot \mid q)}
    \left[
    \frac{
    \sum_{i=1}^{G}\sum_{t=1}^{|o_i|}
    \rho_{i,t}(\theta)
    \hat{R}_{i,t}^{\lambda}(\theta)
    M_{i,t}^{(k)}
    }{
    \sum_{i=1}^{G}\sum_{t=1}^{|o_i|}
    M_{i,t}^{(k)}
    }
    \right].
\end{split}
\end{equation}
Here, $M_{i,t}^{(k)}$ determines which token credits are accepted at training step $k$. By changing this mask over time, \alg uses different allocation rules for exploration and refinement.

\vspace{-7.5pt}
\paragraph{Exploration phase.}
The goal of the early phase is to preserve exploration while avoiding destructive teacher corrections. For $k<T_{\text{switch}}$, \alg keeps dense supervision but filters out excessively negative credits:
\begin{equation}
\label{eq:exploration_mask}
    M_{i,t}^{(k)}
    =
    \mathbb{I}
    \left[
    R_{i,t}(\theta)
    \ge
    \frac{\log \lambda}{1-\lambda}
    \right].
\end{equation}
This mask rejects only credits whose negative magnitude exceeds the bounded-credit threshold in Section~\ref{sec:reward_clipping}, delaying overly aggressive correction that could prematurely prune candidate reasoning paths. As shown in \autoref{fig:multi}, the first phase improves both Avg@32 and Pass@32, suggesting that bounded negative correction preserves useful exploration rather than merely slowing optimization.

\vspace{-7.5pt}
\paragraph{Refinement phase.}
The goal of the later phase is to concentrate update budget on unresolved decisions. For $k\ge T_{\text{switch}}$, \alg switches to the entropy-guided allocation rule from Section~\ref{sec:token-level-dynamic-sampling}:
\begin{equation}
\label{eq:refinement_mask}
    M_{i,t}^{(k)}
    =
    \mathbb{I}
    \left[
    H_t^i
    \ge
    \tau_\beta
    \right].
\end{equation}
This mask selects high-entropy positions where the student remains uncertain and teacher feedback is more actionable. Thus, the schedule first preserves exploratory support and then refines uncertain token decisions. Empirically, \autoref{fig:multi} shows that this transition further improves Avg@32 while maintaining the diversity gains reflected by Pass@32.

%% file: figures/entropy.tex

\begin{figure*}[t]
\centering
\small
\begin{minipage}[t]{0.49\linewidth}
\vspace{0pt}
\centering
\small
\includegraphics[width=1.0\linewidth]{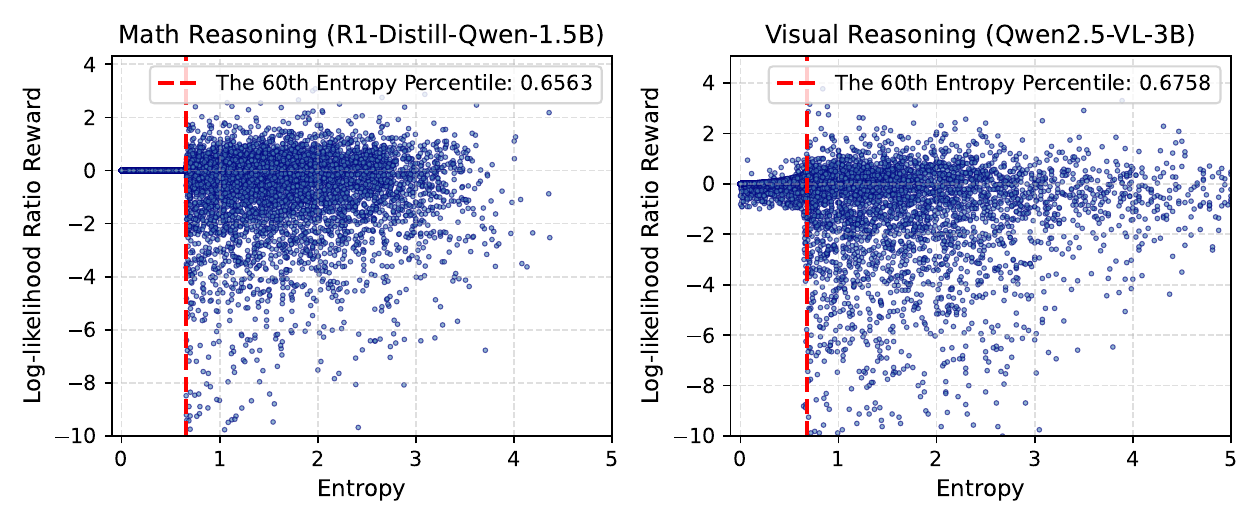}
\caption{\textbf{Correlation between token entropy and credit.} Experimental results on math reasoning and visual reasoning benchmarks demonstrate that credits in the bottom 60th entropy percentile are heavily concentrated around zero. This suggests that while teacher and student policies may diverge overall, they remain highly consistent on low-entropy, deterministic tokens, with significant deviations occurring primarily in high-entropy regimes.}
\label{fig:entropy}
\end{minipage}
\hfill
\begin{minipage}[t]{0.49\linewidth}
\vspace{0pt}
\centering
\small
\includegraphics[width=1.0\linewidth]{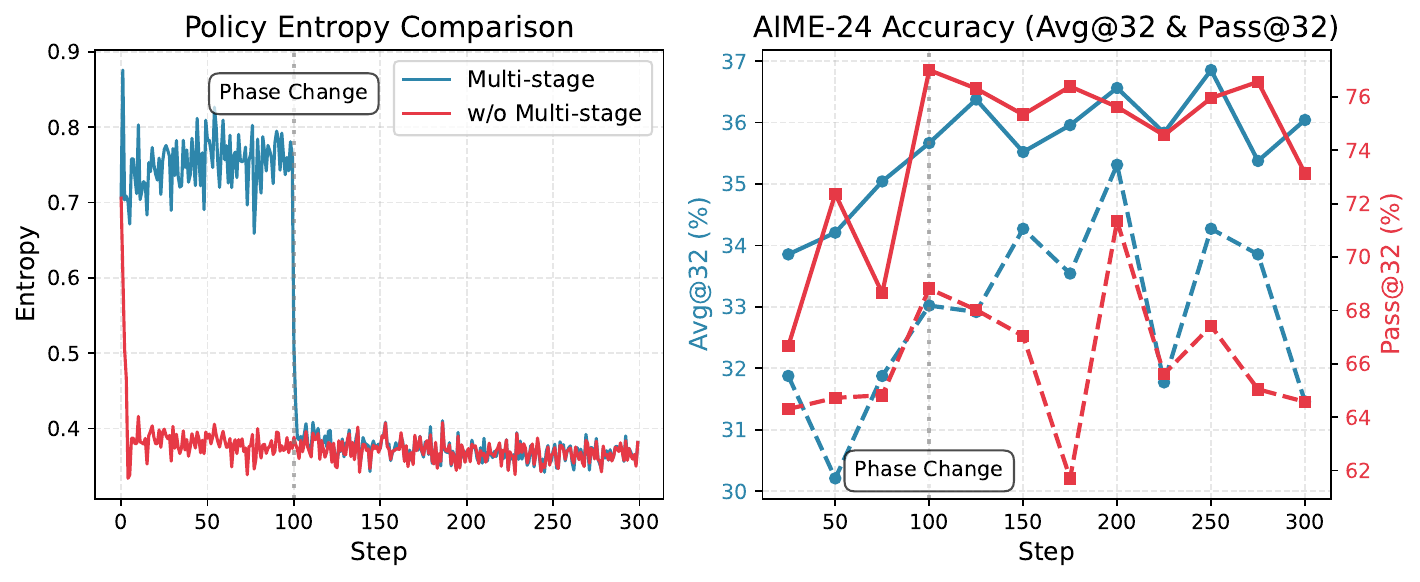}
\caption{\textbf{Analysis of exploration-to-refinement scheduling.} \textbf{(Left)} The proposed multi-stage approach prevents entropy collapse and preserves diverse modes during the initial exploration phase by masking strongly negative credits, whereas the baseline suffers from rapid entropy collapse. \textbf{(Right)} The multi-stage strategy (solid lines) achieves superior final performance to the baseline (dashed lines) in both quality (i.e., Avg@32) and diversity (i.e., Pass@32).}
\label{fig:multi}
\end{minipage}
\vspace{-7.5pt}
\end{figure*}

%% file: sections/experiments.tex
\vspace{-7.5pt}
\input{tables/main}
\section{Experimental Results}\label{sec:exp}

\subsection{Main Results: Math Reasoning}\label{sec:math}
\paragraph{Setup.} We conduct on-policy distillation on DeepSeek-R1-Distill-Qwen-1.5B and 7B \citep{guo2025deepseek}, employing SkyWork-OR1-7B (or Phi-4-mini-reasoning \cite{xu2025phi}) and SkyWork-OR1-32B-Preview \citep{he2025skywork} as teachers, respectively. For training, we utilize the dataset proposed by \citet{yan2025learning}, which contains 45k prompts. While all 1.5B models in \autoref{tab:math} are trained for 300 steps for fair comparison, we extend the training of \alg to 600 steps for the sample efficiency analysis in \autoref{fig:teaser}(a). Detailed training setup is described in Appendix~\ref{app:training}.

\vspace{-7.5pt}
\paragraph{Evaluation.} We evaluate all models on six competition-level mathematical reasoning benchmarks: AIME-24, AIME-25, AMC-23, MATH-500~\citep{hendrycks2020measuring}, Minerva Math~\citep{lewkowycz2022solving}, and Olympiad Bench~\citep{he-etal-2024-olympiadbench}. For AIME-24, AIME-25, and AMC-23, we report Avg@32 to ensure robust evaluation considering the relatively small test sets. For the remaining three benchmarks, we report \textit{Pass@1}. In all evaluations, we use a temperature of 0.6 and a top-p value of 0.95.

\vspace{-7.5pt}
\paragraph{Result 1: Better training sample-efficiency.} As shown in \autoref{fig:teaser}(a), \alg matches ProRL~\citep{liu2025prorl} in 600 steps (vs. 2000); when normalized by the total number of generated sequences, computed as steps $\times$ batch size $\times$ rollouts, this yields a $>6.7\times$ efficiency gain. It surpasses DeepScaleR-1.5B-Preview~\citep{luo2025deepscaler} and DeepMath-1.5B~\citep{he2025deepmath} even earlier at 300 steps ($>12\times$ efficiency), and notably outperforms standard OPD (300 steps) in just 150 steps. Finally, superior performance over GRPO in our re-implementation under identical conditions (\autoref{tab:math}) confirms that our gains stem from algorithmic efficacy rather than experimental setup.

\vspace{-7.5pt}
\paragraph{Result 2: Robustness to different model family.} Using the sequence alignment \cite{patiño2025unlocking, boizard2025towards}, we distill Phi-4-Mini-Reasoning into DeepSeek-R1-Distill-Qwen-1.5B, where the teacher and student come from different model families and use different vocabularies. This setting induces a large teacher-student distribution mismatch, under which standard OPD suffers severe negative transfer, reducing average performance from 44.4 to 35.1. In contrast, \alg improves the student to 47.0, showing that selective credit allocation can stabilize OPD even when teacher feedback is transferred across mismatched vocabularies and model families.

\input{figures/math_7b}
\paragraph{Result 3: Scaling to large policy models.} We further evaluate \alg on a stronger student, DeepSeek-R1-Distill-Qwen-7B. As shown in \autoref{fig:math-7b}, standard OPD is unstable: performance drops sharply early in training and fails to recover beyond the base model on AIME-24. In contrast, \alg prevents early degradation and steadily improves across benchmarks, demonstrating that selective credit allocation remains effective when scaling OPD to stronger policy models.

\input{tables/visual}
\vspace{-5pt}
\subsection{Main Results: Visual Reasoning}\label{sec:visual}
We present the primary evaluation results on visual reasoning tasks below. Please refer to Appendix~\ref{app:analysis} for additional ablation studies  and detailed discussions.

\vspace{-7.5pt}
\paragraph{Setup.} We adopt Qwen2.5-VL-3/7B-Instruct~\citep{bai2025qwen2} as the student policy and Qwen2.5-VL-32B-Instruct, trained with NoisyRollout \citep{liu2025noisyrollout}, as the teacher. We train the student model on Geometry3K \citep{lu2021inter}, which focuses on geometric problem solving and comprises approximately 2.1K training samples. Following the protocol of \citet{liu2025noisyrollout}, we pre-process this dataset by converting all multiple-choice questions into free-form answer formats to mitigate reward hacking and reduce the likelihood of answer guessing. Detailed training setup is described in Appendix~\ref{app:training}.

\vspace{-7.5pt}
\paragraph{Evaluation.} We assess across six benchmarks: five visual reasoning benchmarks, including the test split of Geometry3K, MathVerse~\citep{zhang2024mathverse}, MathVision~\citep{wang2024measuring}, MathVista~\citep{lu2023mathvista}, and WeMath~\citep{qiao2025we}, as well as one visual perception benchmark, HallusionBench~\citep{guan2024hallusionbench}. By following the evaluation protocol of \citet{liu2025noisyrollout}, we employ greedy sampling and nucleus sampling~\citep{Holtzman2020The} with a temperature of 0.6 and a top-p of 0.95 for model inference and use Gemini-2.0-Flash-001 \citep{team2023gemini} as the judge model to parse generated responses.

\vspace{-7.5pt}
\paragraph{Result 1: Efficacy on compact models.} As shown in \autoref{tab:visual}, \alg achieves superior overall performance compared to standard OPD, GKD \cite{agarwal2024onpolicy}, and RL baselines across visual reasoning and perception benchmarks for both 3B and 7B models. \alg also surpasses specialized perception algorithms like NoisyRollout~(7B) and PAPO~(\cite{wang2025perception}; 3B). The Top-$K$ study in Appendix~\ref{app:analysis} further confirms that \alg extends beyond sampled-token supervision by outperforming Top-5 approximated full-vocabulary OPD methods. This highlights its robustness for compact models: by selectively applying teacher credits, REOPOLD avoids unnecessary imitation that can be harmful in low-capacity regimes. 

\vspace{-7.5pt}
\paragraph{Result 2: Superior test-time scaling.} We evaluate test-time scaling by measuring \textit{Pass@}$K$ as a function of parallel thinking latency which is the average time to generate $K$ responses per question. As shown in \autoref{fig:teaser}(b), \alg achieves stronger scaling curves than standard OPD and the Qwen2.5-VL-32B teacher, reaching up to $3.32\times$ higher inference efficiency in terms of Pass@$K$. This gain comes from combining higher generation quality than standard OPD with the lower latency of a compact student model. Extended Maj@$K$ results in \autoref{fig:latency-extension} show the same trend across test-time scaling metrics.

\input{tables/ablation}

\vspace{-5pt}
\subsection{Analysis: Visual Reasoning}
We conduct additional analyses using the visual reasoning setup in Section~\ref{sec:visual}, unless otherwise specified, to further evaluate the effectiveness and generalizability of \alg.

\vspace{-5pt}
\paragraph{Result 1: Impact of module design.} Table~\ref{tab:visual-ablation} validates the three axes of selective credit allocation: (1) bounded negative credit (2) entropy-guided token selection, and (3) exploration-to-refinement scheduling. Each component improves over standard OPD by addressing a distinct allocation failure: overly large negative credits, redundant low-entropy tokens, and the changing usefulness of teacher feedback over training. Combining all three achieves the best average performance for both 3B and 7B students. Consistently, \autoref{fig:entropy_3b_7b} shows that selecting a smaller fraction of high-entropy tokens (\textit{e.g.,} $\beta=0.2$ or $\beta=0.5$) accelerates convergence and improves final accuracy, supporting the view that filtering redundant low-entropy tokens mitigates gradient dilution. We refer the reader to \autoref{fig:clip_sensitivity} and \autoref{fig:multi} for extended analyses on bounded negative credit and exploration-to-refinement scheduling.

\vspace{-5pt}
\input{tables/ablation2}
\paragraph{Result 2: Scalability with longer training.} We further evaluate scalability by extending training to 1000 steps for both 3B and 7B students. As shown in \autoref{fig:1000_step}, standard OPD does not consistently benefit from additional optimization: it saturates around 300 steps for the 3B model and gradually degrades for the 7B model. In contrast, \alg continues improving beyond the exploration-to-refinement transition at step 200 and maintains higher performance across both model sizes. This suggests that \alg better stabilizes long-horizon OPD, enabling continued policy refinement with longer training. Detailed 300-step results are additionally provided in \autoref{tab:longer-3b} and \autoref{tab:longer-7b}.

\vspace{-5pt}
\paragraph{Result 3: Extension to self-distillation.} We further test whether selective credit allocation extends to on-policy self-distillation (OPSD). In OPSD, the same model serves as both teacher and student: the teacher policy is conditioned on privileged information, such as ground-truth solutions, while the student policy observes only the original prompt~\citep{zhao2026self, shenfeld2026self}. Applying our selective credit allocation strategy to OPSD yields Relaxed OPSD. As shown in \autoref{tab:opsd}, Relaxed OPSD consistently improves average accuracy over OPSD for both Qwen2.5-VL-3B-Instruct (43.03 to 45.62) and Qwen2.5-VL-7B-Instruct (49.67 to 50.54). This shows that selective credit allocation is not limited to external-teacher distillation, but also stabilizes privileged self-distillation.

\input{figures/concurrent}
\vspace{-5pt}
\paragraph{Result 4: Comparison with other OPD objectives.} We further compare \alg with recent OPD objectives. Revisiting-OPD matches the teacher's top-$K$ local support~\citep{fu2026revisiting}, while the overlap-restricted variant uses the shared-support objective from~\citet{li2026rethinking}. As shown in Figure~\ref{fig:reb-concurrent}, \alg achieves the highest average accuracy of 52.18, outperforming Revisiting-OPD (51.20) and the overlap-restricted variant (51.08) by 0.98 and 1.10 points, respectively. These results suggest that selective credit allocation provides benefits beyond modifying the supervision support alone.

\vspace{-5pt}
\paragraph{Result 5: Extension to Qwen3-VL.} We additionally evaluate \alg on Qwen3-VL~\citep{yang2025qwen3} to examine its effectiveness across different model architectures. As shown in Figure~\ref{fig:reb-qwen3}, \alg consistently outperforms standard OPD across all six benchmarks, improving average accuracy from 56.35 to 58.48 (+2.13 points). This demonstrates that the benefits of selective credit allocation extend beyond Qwen2.5-VL.

\vspace{-5pt}
\subsection{Extension: Agentic Reasoning with Visual Tool-Use}\label{sec:tool}
Traditional visual reasoning approaches typically treat images as static inputs, limiting the model's ability to actively explore visual information. To address this limitation, we implement image operation tools that enable agents to zoom into specific regions, select key frames, and perform other visual manipulations. This approach, following Pixel-Reasoner~\citep{su2025pixel}, enhances reasoning capabilities over dense visual data.

\input{tables/pixel}
\vspace{-7.5pt}
\paragraph{Setup.} We implement our proposed method based on the VerlTool~\citep{jiang2025verltool} framework. We adopt the SFTed Qwen2.5-VL-3B-Instruct~\citep{jiang2025verltool} as the student policy and Pixel-Reasoner-7B~\citep{su2025pixel} as the teacher. We use the official training dataset from Pixel-Reasoner, comprising 15K queries from InfographicVQA, supplemented by additional public datasets. Detailed training setup is described in Appendix~\ref{app:training}.

\vspace{-7.5pt}
\paragraph{Evaluation.} Following \citet{su2025pixel}, we evaluate our model and baselines on four representative multi-modal benchmarks using nucleus sampling with temperature of 1.0 and a top-p of 1.0: test-split of Pixel-Reasoner, V-Star~\citep{wu2024v}, InfographicVQA~\citep{mathew2022infographicvqa}, and TallyQA~\citep{acharya2019tallyqa}. This selection offers a wide spectrum of visual understanding tasks, ranging from fine-grained object recognition to high-level reasoning in both static and dynamic scenarios.

\vspace{-7.5pt}
\paragraph{Results.} As reported in Table \ref{tab:pixel}, \alg outperforms both standard OPD and GRPO, notably surpassing the latter even when it utilizes the complex reward designs proposed by \citet{su2025pixel}. Although GRPO achieves slightly higher accuracy on the Pixel test split, \alg demonstrates superior performance across other benchmarks, indicating stronger generalization capabilities. Furthermore, we show in \autoref{fig:pixel} that \alg exhibits better sample efficiency compared to standard OPD and GRPO. Unlike traditional RL approaches that necessitate intricate reward engineering for sophisticated agentic tasks, \alg can be applied directly.

%% file: tables/main.tex
\begin{table*}[t]
\centering
\caption{Performance comparison of \alg and distillation baselines (SFT and standard OPD) on mathematical reasoning benchmarks across different teacher models. Accuracy (\%) is reported for all benchmarks. The best result in each column is shown in \textbf{bold}, and the second-best is \underline{underlined}. The results are reported as mean $\pm$ standard deviation over 5 runs.}
\resizebox{1.0\linewidth}{!}{
\begin{tabular}{lccccccc}
\toprule[0.1em]
\textbf{Model} & \textbf{AIME-24} & \textbf{AIME-25} & \textbf{AMC-23} & \textbf{MATH-500} & \textbf{Minerva Math} & \textbf{Olympiad Bench} & \textbf{AVG.}  \\
\midrule
\rowcolor{blue!10} \multicolumn{8}{c}{\textbf{SkyWork-OR1-Math-7B \citep{he2025skywork} $\rightarrow$ DeepSeek-R1-Distill-Qwen-1.5B \citep{guo2025deepseek}}} \\
\midrule
SkyWork-OR1-Math-7B & 69.8 & 52.3 & 94.1 & 95.8 & 49.3 & 73.5 & 72.5  \\
\midrule
R1-Distill-Qwen-1.5B & 28.6 & 22.7 & 62.6 & 82.9 & 26.4 & 43.6 & 44.4  \\
\textcolor{black!90}{\;+ GRPO} & \textcolor{black!90}{31.8}$_{\pm 1.0}$ & \textcolor{black!90}{23.7}$_{\pm 0.9}$ & \textcolor{black!90}{62.0}$_{\pm 1.1}$ & \textcolor{black!90}{85.4}$_{\pm 0.6}$ & \textcolor{black!90}{33.8}$_{\pm 0.3}$ & \textcolor{black!90}{49.8}$_{\pm 0.6}$ & \textcolor{black!90}{47.8}$_{\pm 0.3}$ \\
\;+ SFT & 33.5$_{\pm 1.5}$ & 24.6$_{\pm 1.9}$ & 76.1$_{\pm 1.2}$ & 86.6$_{\pm 0.8}$ & \underline{36.4}$_{\pm 0.8}$ & 55.6$_{\pm 0.8}$ & 51.5$_{\pm 0.9}$ \\
\;+ Standard OPD & \underline{37.1}$_{\pm 1.4}$ & \underline{30.6}$_{\pm 0.6}$ & \underline{80.2}$_{\pm 0.9}$ & \underline{88.0}$_{\pm 0.5}$ & 34.6$_{\pm 0.4}$ & \underline{56.0}$_{\pm 0.8}$ & \underline{54.4}$_{\pm 0.2}$  \\
\;+ \alg & \textbf{41.6}$_{\pm 1.1}$ & \textbf{32.6}$_{\pm 0.4}$ & \textbf{83.0}$_{\pm 1.2}$ & \textbf{89.2}$_{\pm 0.4}$ & \textbf{38.6}$_{\pm 0.7}$ & \textbf{57.3}$_{\pm 0.5}$ & \textbf{57.1}$_{\pm 0.3}$ \\
\midrule
\rowcolor{blue!10} \multicolumn{8}{c}{\textbf{Phi-4-Mini-Reasoning \citep{xu2025phi} $\rightarrow$ DeepSeek-R1-Distill-Qwen-1.5B \citep{guo2025deepseek}}} \\
\midrule
Phi4-Mini-Reasoning & 55.6 & 41.4 & 91.8 & 94.6 & 39.1 & 64.5 & 64.5 \\
\midrule
\;+ Standard OPD & 13.3$_{\pm 0.5}$ & 21.0$_{\pm 1.1}$ & 45.5$_{\pm 2.1}$ & 78.6$_{\pm 2.7}$ & 19.4$_{\pm 1.4}$ & 32.8$_{\pm 1.0}$ & 35.1$_{\pm 1.1}$ \\
\;+ \alg & \textbf{32.1}$_{\pm 1.7}$ & 24.6$_{\pm 0.7}$ & 65.2$_{\pm 1.2}$ & 83.7$_{\pm 1.3}$ & 30.2$_{\pm 0.5}$ & 45.9$_{\pm 0.6}$ & 47.0$_{\pm 0.4}$ \\
\bottomrule[0.1em]
\end{tabular}}\label{tab:math}
\end{table*}

%% file: figures/math_7b.tex
    

\begin{wrapfigure}{r}{0.5\textwidth}
\vspace{-15pt}
    \centering
    \includegraphics[width=1.0\linewidth]{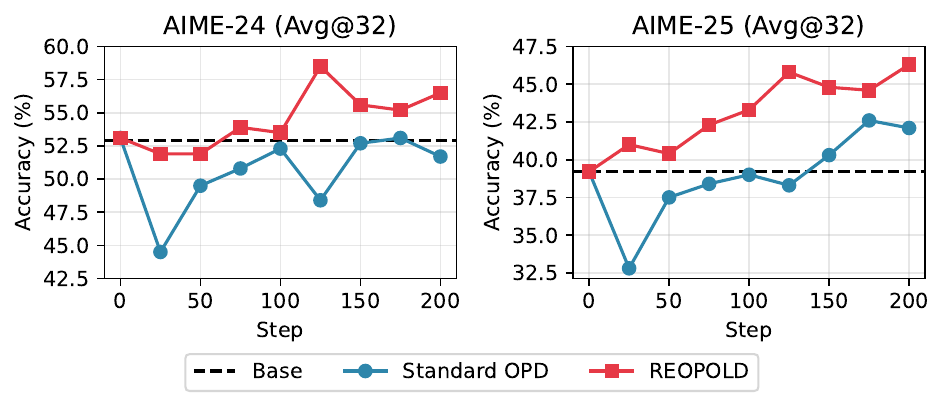}
    \caption{
        Scaling to a 7B student exposes instability in standard OPD, while \alg maintains stable and consistent improvement.
    }
    \vspace{-10pt}
    \label{fig:math-7b}
\end{wrapfigure}

%% file: tables/visual.tex
\begin{table*}[t]
\centering
\small
\caption{Performance comparison of vision-language models on visual reasoning and perception benchmarks. Accuracy (\%) is reported for all benchmarks. The best result in each column is shown in \textbf{bold}, and the second-best is \underline{underlined}. $^\ddagger$ denotes results obtained with the Top-5 approximation, as the full-vocabulary implementation incurs memory issues. Our reproduced OPD results are reported as mean $\pm$ standard deviation over 5 runs.}
\resizebox{1.0\linewidth}{!}{
\begin{tabular}{lccccccc}
\toprule[0.1em]
\textbf{Model} & \textbf{Geo3K} & \textbf{MathVerse} & \textbf{MathVision} & \textbf{MathVista} & \textbf{WeMath} & \textbf{Hallusion} & \textbf{AVG.} \\
\midrule
Qwen2.5-VL-32B-Instruct\,+\,NoisyRollout & 56.74 & 58.86 & 39.82 & 78.30 & 75.51 & 72.45 & 63.61 \\
\midrule
\rowcolor{blue!10} \multicolumn{8}{c}{\textbf{Qwen2.5-VL-32B-Instruct + NoisyRollout \citep{liu2025noisyrollout} $\rightarrow$ Qwen2.5-VL-3B-Instruct}} \\
\midrule
Qwen2.5-VL-3B-Instruct & 26.46$_{\pm 1.58}$  & 35.58$_{\pm 0.44}$  & 22.83$_{\pm 0.66}$  & 59.40$_{\pm 0.49}$  & 53.41$_{\pm 0.41}$  & {61.51}$_{\pm 0.39}$ & 43.20$_{\pm 0.54}$  \\
\textcolor{gray!90}{\;+ PAPO}\,\citep{wang2025perception} & \textcolor{gray!90}{32.95} & \textcolor{gray!90}{\underline{40.65}} & \textcolor{gray!90}{24.16} & \textcolor{gray!90}{\textbf{65.10}} & \textcolor{gray!90}{63.62} & \textcolor{gray!90}{\underline{61.62}} & \textcolor{gray!90}{\underline{48.02}} \\
\;+ Standard OPD \cite{lu2025onpolicy} & \underline{48.09}$_{\pm 0.39}$   & 33.25$_{\pm 0.41}$   & \underline{24.78}$_{\pm 0.62}$  & \underline{62.50}$_{\pm 0.56}$  & \underline{64.48}$_{\pm 0.43}$  & 60.83$_{\pm 0.38}$  & 48.99$_{\pm 0.34}$   \\
\;+ \alg & \textbf{50.58}$_{\pm 0.64}$  & \textbf{46.40}$_{\pm 0.40}$  & \textbf{26.39}$_{\pm 0.38}$  & {61.50}$_{\pm 0.45}$   & \textbf{64.60}$_{\pm 0.23}$   & \textbf{63.62}$_{\pm 0.21}$   & \textbf{52.18}$_{\pm 0.21}$   \\
\midrule
\rowcolor{blue!10} \multicolumn{8}{c}{\textbf{Qwen2.5-VL-32B-Instruct + NoisyRollout \citep{liu2025noisyrollout} $\rightarrow$ Qwen2.5-VL-7B-Instruct}} \\
\midrule
Qwen2.5-VL-7B-Instruct & 39.77$_{\pm 1.33}$  & 45.72$_{\pm 0.69}$  & 25.05$_{\pm 0.68}$  & 67.80$_{\pm 0.45}$  & 64.77$_{\pm 0.53}$  & 65.62$_{\pm 1.39}$  & 51.46$_{\pm 0.13}$  \\
\textcolor{gray!90}{\;+ GRPO}\,\citep{shao2024deepseekmath} & \textcolor{gray!90}{51.4} & \textcolor{gray!90}{50.8} & \textcolor{gray!90}{27.3} & \textcolor{gray!90}{70.5} & \textcolor{gray!90}{67.4} & \textcolor{gray!90}{69.8} & \textcolor{gray!90}{56.20} \\
\textcolor{gray!90}{\;+ NoisyRollout}\,\citep{liu2025noisyrollout} & \textcolor{gray!90}{50.08} & \textcolor{gray!90}{\textbf{53.14}} & \textcolor{gray!90}{26.64} & \textcolor{gray!90}{\underline{72.00}} & \textcolor{gray!90}{\textbf{70.57}} & \textcolor{gray!90}{\textbf{70.66}} & \textcolor{gray!90}{\underline{57.18}} \\
\;+ Standard OPD \cite{lu2025onpolicy} & \underline{51.75}$_{\pm 0.40}$   & 47.71$_{\pm 0.61}$   & \underline{28.79}$_{\pm 0.32}$  & 71.27$_{\pm 0.68}$  & \underline{70.06}$_{\pm 0.64}$   & 69.51$_{\pm 1.29}$   & 56.51$_{\pm 0.19}$   \\
\;+ GKD$^\ddagger$ \cite{agarwal2024onpolicy} & 52.07$_{\pm 0.84}$  & 48.31$_{\pm 0.47}$  & 27.98$_{\pm 0.53}$  & 70.89$_{\pm 0.36}$  & 69.34$_{\pm 0.41}$  & 69.75$_{\pm 1.13}$  & 56.39$_{\pm 0.24}$  \\
\;+ \alg & \textbf{53.58}$_{\pm 0.71}$ & \underline{51.43}$_{\pm 0.46}$ & \textbf{29.21}$_{\pm 0.49}$  & \textbf{72.40}$_{\pm 0.68}$ & 69.77$_{\pm 0.24}$  & \underline{70.14}$_{\pm 0.86}$ & \textbf{57.76}$_{\pm 0.31}$ \\
\bottomrule[0.1em]
\end{tabular}}\label{tab:visual}
\end{table*}

%% file: tables/ablation.tex
\begin{figure}[t]
\centering
\small
\begin{minipage}[t]{0.51\columnwidth}
\vspace{0pt}
\captionsetup{type=table}
\caption{Performance comparison of Qwen2.5-VL-3/7B-Instruct trained with 32B teacher models on visual reasoning benchmarks. Accuracy (\%) is reported for all benchmarks.}
\label{tab:visual-ablation}
\resizebox{\linewidth}{!}{
\begin{tabular}{l@{\quad}ccc@{\quad}cccc}
\toprule[0.1em]
 & (1) & (2) & (3) & \textbf{Geo3K} & \textbf{Verse} & \textbf{Vision} & \textbf{AVG.} \\
\midrule
\rowcolor{blue!10} \multicolumn{8}{c}{\textbf{Qwen2.5-VL-3B-Instruct}}  \\
\midrule
Standard OPD    &        & & & 48.09 & 33.25 & 24.78 & 35.37 \\ \midrule
                & \cmark & & & 48.59 & 45.53 & 25.07 & 39.73  \\
                & \cmark & \cmark & & 50.08 & 45.94 & 26.42 & 40.81 \\ \midrule
\alg            & \cmark & \cmark & \cmark & 50.58 & 46.40 & 26.39 & 41.12  \\
\midrule
\rowcolor{blue!10} \multicolumn{8}{c}{\textbf{Qwen2.5-VL-7B-Instruct}}  \\
\midrule
Standard OPD     & & & & 51.75 & 47.71 & 28.79 & 42.75  \\ \midrule
        & \cmark & & & 51.75 & 51.45 & 29.05 & 44.08  \\
        & \cmark & \cmark & & 52.75 & 51.42 & 29.10 & 44.42  \\ \midrule
\alg    & \cmark & \cmark & \cmark & 53.58 & 51.43 & 29.21 & 44.74  \\
\bottomrule[0.1em]
\end{tabular}}
\end{minipage}
\hfill
\begin{minipage}[t]{0.46\textwidth}
\vspace{0pt}
\centering
\small
\includegraphics[width=\textwidth]{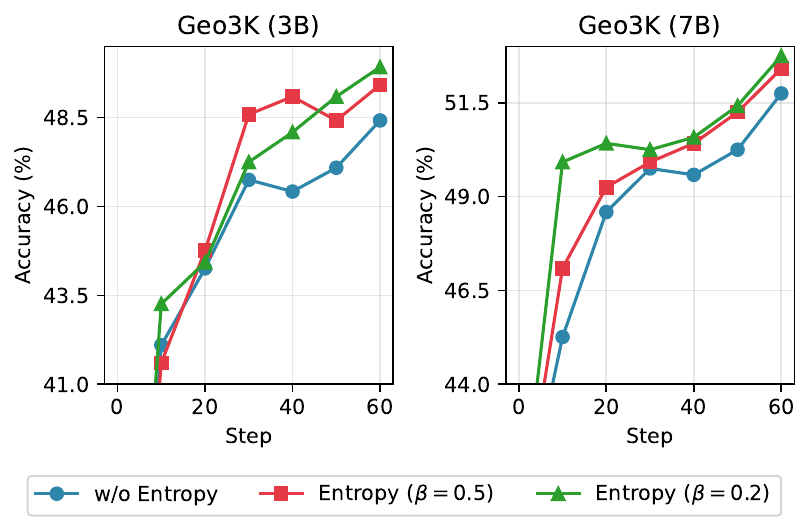}
\caption{Comparison among \alg using different entropy-selection fractions. Our token selection accelerates convergence and leads to superior final accuracy.}
\label{fig:entropy_3b_7b}
\end{minipage}
\vspace{-5pt}
\end{figure}

%% file: tables/ablation2.tex
\begin{figure}[t]
\centering
\small
\begin{minipage}[t]{0.46\textwidth}
\vspace{0pt}
\centering
\small
\includegraphics[width=\textwidth]{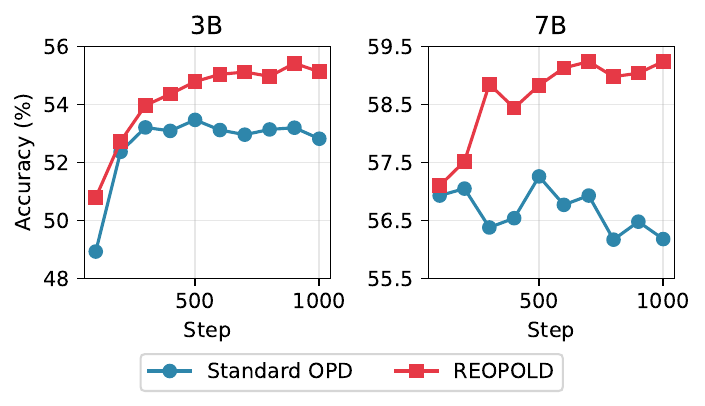}
\caption{Convergence curves of OPD and \alg over 1000 training steps.}
\label{fig:1000_step}
\end{minipage}
\hfill
\begin{minipage}[t]{0.51\columnwidth}
\vspace{0pt}
\captionsetup{type=table}
\caption{Performance comparison of Qwen2.5-VL-3/7B-Instruct trained by OPSD \citep{zhao2026self} on visual reasoning benchmarks. Accuracy (\%) is reported for all benchmarks.}
\label{tab:opsd}
\resizebox{\linewidth}{!}{
\begin{tabular}{lccccc}
\toprule[0.1em]
 & \textbf{Geo3K} & \textbf{Verse} & \textbf{Vision} & \textbf{Vista} & \textbf{AVG.} \\
\midrule
\rowcolor{blue!10} \multicolumn{6}{c}{\textbf{Qwen2.5-VL-3B-Instruct}}  \\
\midrule
OPSD            & 44.27 & 41.94 & 24.16 & 61.75 & 43.03  \\ 
Relaxed OPSD    & \textbf{49.82} & \textbf{45.14} & \textbf{25.71} & \textbf{61.82} & \textbf{45.62}  \\ 
\midrule
\rowcolor{blue!10} \multicolumn{6}{c}{\textbf{Qwen2.5-VL-7B-Instruct}}  \\
\midrule
OPSD            & 50.72 & 49.41 & 28.22 & 70.33 & 49.67  \\ 
Relaxed OPSD    & \textbf{51.82} & \textbf{50.43} & \textbf{29.07} & \textbf{70.84} & \textbf{50.54}  \\ 
\bottomrule[0.1em]
\end{tabular}}
\end{minipage}
\end{figure}

%% file: figures/concurrent.tex
\begin{figure}[t]
\centering
\begin{minipage}[t]{0.49\linewidth}
\centering
\includegraphics[width=\linewidth]{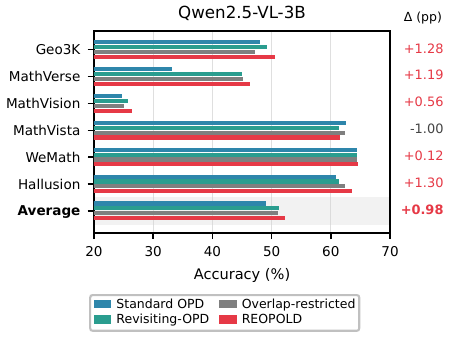}
\caption{OPD comparisons on Qwen2.5-VL-3B. The axis starts at 20; $\Delta$ is \alg minus the best other method in each row (pp).}
\label{fig:reb-concurrent}
\end{minipage}\hfill
\begin{minipage}[t]{0.49\linewidth}
\centering
\includegraphics[width=\linewidth]{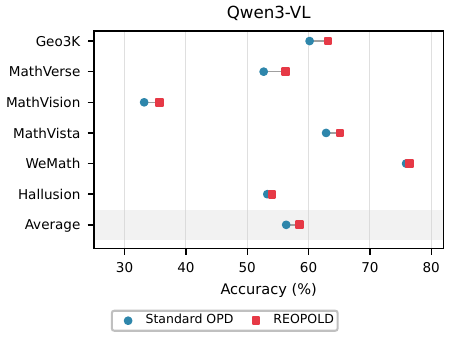}
\caption{Extension to Qwen3-VL. Paired markers compare standard OPD and \alg across six benchmarks. Shading marks the average.}
\label{fig:reb-qwen3}
\end{minipage}
\vspace{-5pt}
\end{figure}

%% file: tables/pixel.tex
\begin{figure*}[t]
\centering
\small
\begin{minipage}[b]{0.71\textwidth}
\centering
\captionof{table}{Performance comparison of vision-language models on agentic visual tool-use tasks. Accuracy (\%) is reported for all benchmarks. The best and the second-best results in each column are shown in \textbf{bold} and \underline{underlined}, respectively.}
\resizebox{1.0\textwidth}{!}{
\addtolength{\tabcolsep}{0.5pt}
\begin{tabular}{lccccc}
\toprule[0.1em]
\textbf{Model} & \textbf{Pixel} & \textbf{V-Star} & \textbf{InfoVQA}  & \textbf{TallyQA} & \textbf{AVG.} \\
\midrule
Pixel-Reasoner-7B  & 64.00 & 84.29 & 74.37 & 73.69 & 74.05 \\
\midrule
\rowcolor{blue!10} \multicolumn{6}{c}{\textbf{Pixel-Reasoner-7B \citep{su2025pixel} $\rightarrow$ Qwen2.5-VL-3B-Instruct + SFT \citep{jiang2025verltool}}} \\
\midrule
Qwen2.5-VL-3B-Instruct + SFT & 46.00 & 71.20 & 34.88 & 56.91 & 52.25 \\
\textcolor{black!90}{\;+ GRPO}~\citep{jiang2025verltool} & \textcolor{black!90}{\textbf{60.00}} &  \textcolor{black!90}{\underline{76.96}} & \textcolor{black!90}{{59.47}}  & \textcolor{black!90}{60.56} & \textcolor{black!90}{\underline{64.25}} \\
\;+ Standard OPD & 52.00 & 76.55 & \underline{61.09} & \underline{64.34} & 63.27 \\
\;+ \alg & \underline{57.00} & \textbf{77.43} & \textbf{63.12} & \textbf{65.43} & \textbf{65.75} \\
\bottomrule[0.1em]
\end{tabular}
}\label{tab:pixel}
\end{minipage}
\hfill
\begin{minipage}[t]{0.27\textwidth}
\vspace{-93pt}
\centering
\small
\includegraphics[width=\textwidth]{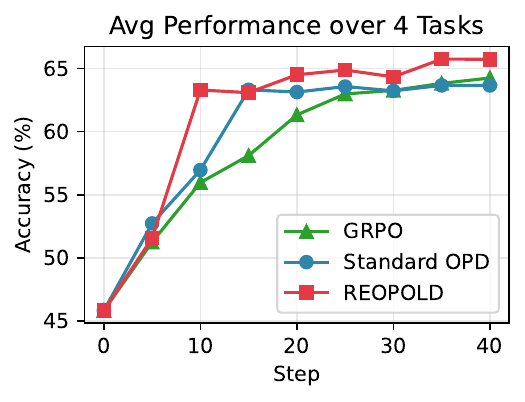}
\caption{\textbf{Average score by training step.} \alg outperforms all baselines at 50\% training.}
\label{fig:pixel}
\end{minipage}
\vspace{-7.5pt}
\end{figure*}

%% file: sections/conclusion.tex
\vspace{-5pt}
\section{Conclusion}
\vspace{-5pt}
We studied sampled-token OPD through the lens of selective credit allocation. Standard OPD applies teacher credits uniformly despite their imbalanced structure, causing inefficient updates and premature suppression of useful reasoning paths. To address this, we introduced \alg, a unified framework that regulates teacher feedback along three axes: entropy-guided token selection focuses supervision on uncertain decisions, bounded negative credit controls update magnitude, and an exploration-to-refinement schedule changes the allocation rule over training. Empirically, \alg improves OPD across mathematical, visual, and agentic tool-use reasoning, and remains effective under cross-vocabulary distillation, self-distillation, longer training, and test-time scaling. These results suggest selective credit allocation is a practical principle for robust reasoning distillation in compact models.

%% file: sections/related-work-extension.tex
\clearpage
\section{Related Work}\label{app:related}

\paragraph{Reasoning models.} Reasoning models represent a distinct class of machine learning systems designed to execute structured, logical, and multi-step inference over input queries~\cite{creswell2022selection, huang-chang-2023-towards, chen2025towards}. In contrast to standard predictive models that rely on direct input-to-output mapping, reasoning models emulate human cognitive processes by integrating learned knowledge with stepwise deduction~\cite{creswell2022selection}, chain-of-thought processing~\cite{wei2022chain}, or symbolic manipulation~\cite{weng2024mastering}. These capabilities are essential for tasks involving complex problem-solving~\cite{wang2023selfconsistency,giadikiaroglou-etal-2024-puzzle}, question answering~\cite{zhang2024chain, santana2025question}, planning~\cite{yao2023tree,hao-etal-2023-reasoning}, and commonsense inference~\cite{rajani-etal-2019-explain,zhao2023large}, where a single forward pass is often insufficient. Recent advancements have increasingly integrated reasoning into LLMs and neuro-symbolic architectures~\cite{li2024neuralsymbolic,calanzone2025logically}. This integration allows models to decompose complex problems into intermediate steps, verify logical consistency~\cite{calanzone2025logically}, and generate interpretable solutions~\cite{barbiero2023interpretable,yao2023react}. Fundamentally, reasoning models ensure that outputs are both accurate and justifiable, emphasizing the rationale behind a decision as much as the decision itself.

\paragraph{Policy optimization for reasoning models.} Recent advancements in policy optimization focus on enhancing the sample efficiency, stability, and reasoning depth of LLMs. While initial approaches relied on standard outcome-based RL, recent works demonstrate that scaling RL on smaller architectures, as seen in DeepScaleR~\citep{luo2025deepscaler} and Skywork OpenReasoner~\citep{he2025skywork}, can achieve performance rivaling proprietary frontiers like OpenAI-o1. To improve algorithmic stability beyond basic group-relative updates, GSPO~\citep{zheng2025group} introduces step-level granularity for precise credit assignment, whereas GMPO~\citep{zhao2025geometric} adopts a group-wise minimax formulation to bolster robustness against distribution shifts. Extending this to hybrid training objectives, KDRL~\cite{xu2025kdrl} proposes a unified framework that synergizes knowledge distillation with RL, effectively balancing teacher supervision and self-exploration. Addressing the critical balance between exploration and exploitation, the Entropy Mechanism~\citep{cui2025entropy} dynamically regulates policy entropy to prevent premature convergence, while LUFFY~\citep{yan2025learning} improve optimization efficiency by effectively leveraging diverse, off-policy trajectories. Furthermore, emphasizing the generation of extended reasoning chains, ProRL~\citep{liu2025prorl} explicitly incentivizes prolonged thought process to expand the models' reasoning boundaries, a capability that underpins state-of-the-art large-scale systems such as MiniMax-M1~\citep{chen2025minimax}.

\paragraph{On-policy distillation for reasoning models.} Traditional knowledge distillation typically relies on offline datasets generated by a teacher model, which creates a distribution mismatch as the student's policy drifts from the static training data \citep{ho-etal-2023-large, hsieh-etal-2023-distilling, ko2025flexjudge}. To bridge this gap, on-policy distillation aligns the student with the teacher's distribution by training on trajectories sampled directly from the student's current policy \citep{gu2024minillm, agarwal2024onpolicy, ko2024distillm}. This paradigm is particularly critical for reasoning tasks, where models must learn to recover from their own logical errors rather than merely mimicking perfect teacher paths~\citep{lu2025onpolicy, patiño2025unlocking}. Recently, there has been a growing movement to adapt on-policy distillation for reasoning tasks. Frontier models like Qwen3~\citep{yang2025qwen3} utilize iterative on-policy feedback to refine long-chain reasoning capabilities, while MiMo-V2-Flash~\citep{xiao2026mimo} demonstrates that such methods achieve superior compute-efficiency by targeting "hard" examples where the student's confidence diverges from the teacher. Furthermore, self-distillation \cite{zhao2026self, shenfeld2026self}, where the same model serves as both teacher and student and the teacher policy is conditioned on privileged information such as the ground-truth solution, has recently gained popularity.

%% file: sections/proof.tex
\clearpage
\section{Mathematical Derivations}

\subsection{A Fixed-credit View}\label{app:proof}
The teacher-derived token credit $R_{i,t}(\theta)$ depends on $\theta$, since it contains the student probability in the denominator. Following the stop-gradient formulation, we treat this credit as a fixed teacher-derived signal during each update. Let $\mathcal{J}^{\mathrm{sg}}_{\mathrm{RKL}}$ denote the corresponding objective. Then, the expected gradient satisfies
\begin{equation}
\begin{split}
    \nabla_\theta \mathcal{J}_{\mathrm{RKL}}(\theta)
    =
    \nabla_\theta \mathcal{J}^{\mathrm{sg}}_{\mathrm{RKL}}(\theta)
    =
    \mathbb{E}
    \left[
    \frac{1}{\sum_i |o_i|}
    \sum_{i,t}
    \rho_{i,t}(\theta)
    R_{i,t}(\theta)
    \nabla_\theta \log \pi_\theta(o_{i,t}|q,o_{i,<t})
    \right].
\end{split}
\end{equation}

We derive the gradient of RKL. From Eq.~\ref{eq:rkl_off}, the RKL objective and its gradient are:
\begin{equation}
    \mathcal{J}_{\text{RKL}}(\theta) = \mathbb{E}_{q \sim Q, o \sim \pi_{\theta_{\text{old}}}(\cdot|q)} \left[ \frac{1}{|o|} \sum_{t=1}^{|o|} \rho_t(\theta) R_t(\theta) \right],
\end{equation}
where $\rho_t (\theta) = \frac{\pi_\theta(o_t | q, o_{<t})}{\pi_{\theta_{\text{old}}} (o_t | q, o_{<t})}$ and $R_t(\theta)= \log \frac{\pi_T(o_t | q, o_{<t})}{\pi_{\theta} (o_t | q, o_{<t})}$. Then following holds,
\begin{align*}
    \nabla_\theta \mathcal{J}_{\text{RKL}}(\theta) &= \mathbb{E}_{q \sim Q, o \sim \pi_{\theta_{\text{old}}}(\cdot|q)} \left[ \frac{1}{|o|} \sum_{t=1}^{|o|} \rho_t(\theta) \nabla_\theta R_t(\theta) +  R_t(\theta) \nabla_\theta \rho_t(\theta)\right], \\
    &= \mathbb{E}_{q \sim Q, o \sim \pi_{\theta_{\text{old}}}(\cdot|q)} \left[ \frac{1}{|o|} \sum_{t=1}^{|o|} \rho_t(\theta) R_t(\theta) \nabla_\theta \log \pi_\theta(o_t | q, o_{<t}) \right] -  \mathbb{E}_{q \sim Q, o \sim \pi_{\theta_{\text{old}}}(\cdot|q)} \left[  \frac{1}{|o|} \sum_{t=1}^{|o|} \frac{\nabla_\theta \pi_\theta(o_t | q, o_{<t})}{\pi_{\theta_{\text{old}}} (o_t | q, o_{<t})} \right], \\
    &= \mathbb{E}_{q \sim Q, o \sim \pi_{\theta_{\text{old}}}(\cdot|q)} \left[ \frac{1}{|o|} \sum_{t=1}^{|o|} \rho_t(\theta) R_t(\theta) \nabla_\theta \log \pi_\theta(o_t | q, o_{<t}) \right] -  \nabla_\theta \mathbb{E}_{q \sim Q, o \sim \pi_{\theta_{\text{old}}}(\cdot|q)} \left[  \frac{1}{|o|} \sum_{t=1}^{|o|} \frac{\pi_\theta(o_t | q, o_{<t})}{\pi_{\theta_{\text{old}}} (o_t | q, o_{<t})} \right], \\
    &= \mathbb{E}_{q \sim Q, o \sim \pi_{\theta_{\text{old}}}(\cdot|q)} \left[ \frac{1}{|o|} \sum_{t=1}^{|o|} \rho_t(\theta) R_t(\theta) \nabla_\theta \log \pi_\theta(o_t | q, o_{<t}) \right] \\
    &-  \nabla_\theta \mathbb{E}_{q \sim Q, o \sim \pi_{\theta}(\cdot|q)} \left[  \frac{1}{|o|} \sum_{t=1}^{|o|} \frac{\pi_\theta(o_t | q, o_{<t})}{\pi_{\theta_{\text{old}}} (o_t | q, o_{<t})} \cdot \frac{\pi_{\theta_{\text{old}}}(o_t | q, o_{<t})}{\pi_\theta (o_t | q, o_{<t})} \right], \\
    &= \mathbb{E}_{q \sim Q, o \sim \pi_{\theta_{\text{old}}}(\cdot|q)} \left[ \frac{1}{|o|} \sum_{t=1}^{|o|} \rho_t(\theta) R_t(\theta) \nabla_\theta \log \pi_\theta(o_t | q, o_{<t}) \right] -  \nabla_\theta \mathbb{E}_{q \sim Q, o \sim \pi_{\theta}(\cdot|q)} \left[  \frac{1}{|o|} \sum_{t=1}^{|o|} 1 \right], \\
    &= \mathbb{E}_{q \sim Q, o \sim \pi_{\theta_{\text{old}}}(\cdot|q)} \left[ \frac{1}{|o|} \sum_{t=1}^{|o|} \rho_t(\theta) R_t(\theta) \nabla_\theta \log \pi_\theta(o_t | q, o_{<t}) \right] \\
\end{align*}
This justifies treating $R_t$ as a fixed token-level credit within each update. The fixed-credit view separates how teacher credit is computed from how the resulting sampled-token credits are allocated during optimization. Our work focuses on the latter: standard RKL allocates update budget uniformly across sampled tokens, regardless of whether their credits are informative, redundant, or extreme.

\subsection{Logit-Level Update-Energy Proxy}\label{app:energy}
Let $p_\theta(\cdot \mid x)=\pi_\theta(\cdot \mid q,o_{<t})$ denote the student distribution at context $x=(q,o_{<t})$, and let $y=o_t$ be the sampled token. For a sampled-token objective with scalar credit $R_{i,t}(\theta)$, the token-level likelihood-ratio update with respect to the logits $z$ is proportional to
\begin{equation*}
    g_{i,t} = R_{i,t}(\theta)\nabla_z \log p_\theta(y\mid x).
\end{equation*}

For a softmax distribution, we have
\begin{equation*}
    \nabla_z \log p_\theta(y\mid x) = e_y - p_\theta(\cdot \mid x),
\end{equation*}
where $e_y$ is the one-hot vector of the sampled token. Therefore,
\begin{equation*}
    \|g_{i,t}\|_2^2 = R_{i,t}(\theta)^2
\|e_y - p_\theta(\cdot \mid x)\|_2^2.
\end{equation*}

Taking expectation over $y\sim p_\theta(\cdot\mid x)$ gives
\begin{equation*}
    \mathbb{E}_{y\sim p_\theta} \left[\|e_y - p_\theta(\cdot \mid x)\|_2^2 \right] = 1-\|p_\theta(\cdot \mid x)\|_2^2.
\end{equation*}

Thus, the expected squared logit-level update induced by token credit $R_{i,t}(\theta)$ is
\begin{equation*}
    \mathbb{E}_{y\sim p_\theta} \left[\|g_{i,t}\|_2^2 \right] = R_{i,t}(\theta)^2 \left( 1-\|p_\theta(\cdot \mid x)\|_2^2 \right).
\end{equation*}

This motivates the update-energy proxy used in Section~\ref{sec:motivation}:
\begin{equation*}
    E_{i,t}(\theta) = R_{i,t}(\theta)^2 \left( 1-\|\pi_\theta(\cdot \mid q,o_{i,<t})\|_2^2 \right).
\end{equation*}

\subsection{Softmax Score Magnitude}\label{app:softmax_score_identity}
We prove Remark~\ref{prop:softmax_score_entropy}. Let $p=\operatorname{softmax}(z)$ over vocabulary $\mathcal{V}$, and denote by $e_y$ the one-hot vector for token $y$. For each coordinate $v\in\mathcal{V}$,
\begin{equation*}
    \frac{\partial}{\partial z_v}\log p_y
    =
    \mathbb{I}[v=y]-p_v,
\end{equation*}
and therefore $\nabla_z\log p_y=e_y-p$. Taking expectation over $y\sim p$ gives
\begin{align*}
    \mathbb{E}_{y\sim p}\left[\|\nabla_z\log p_y\|_2^2\right]
    &=
    \sum_{y\in\mathcal{V}}p_y\|e_y-p\|_2^2 \\
    &=
    \sum_{y\in\mathcal{V}}p_y\left(1-2p_y+\|p\|_2^2\right) \\
    &=
    1-\|p\|_2^2.
\end{align*}
Finally, since $-\log u\ge 1-u$ for $u\in(0,1]$, we have $-p_v\log p_v\ge p_v(1-p_v)$ for every $v$ with $p_v>0$. Summing over $v$ and using continuity at $p_v=0$ yields
\begin{equation*}
    H(p)
    =
    -\sum_{v\in\mathcal{V}}p_v\log p_v
    \ge
    \sum_{v\in\mathcal{V}}p_v(1-p_v)
    =
    1-\|p\|_2^2.
\end{equation*}

\subsection{Derivation of Bounded Threshold}\label{app:clip_proof}
Here, we derive the relationship between the standard log-likelihood ratio and the convex mixture ratio used to motivate our clipping threshold. Since the logarithm is a concave function, for $\forall \lambda \in [0,1)$, Jensen’s inequality implies:
\begin{equation}\label{eq:mixture}
    (1-\lambda) \cdot \log \pi_T (o_t|q, o_{<t}) + \lambda \cdot \log \pi_{\theta} (o_t|q, o_{<t}) \leq \log \left[(1-\lambda) \cdot \pi_T (o_t|q, o_{<t}) + \lambda \cdot \pi_{\theta} (o_t|q, o_{<t}) \right].
\end{equation}

To isolate the log-likelihood ratio $R_{i,t}(\theta) = \log \frac{\pi_{T} (o_t|q, o_{<t})}{\pi_{\theta} (o_t|q, o_{<t})}$, we subtract $\log \pi_{\theta} (o_t|q, o_{<t})$ from both sides and divide by ($1-\lambda$):
\begin{equation}
    R_{i,t}(\theta) = \log \frac{\pi_T (o_t|q, o_{<t})}{\pi_{\theta} (o_t|q, o_{<t})} \leq
    \frac{1}{1-\lambda} \log \frac{(1-\lambda) \pi_T (o_t|q, o_{<t}) + \lambda \pi_{\theta} (o_t|q, o_{<t})}{\pi_{\theta} (o_t|q, o_{<t})}.
\end{equation}
This inequality upper-bounds the log-ratio between the teacher and student policies by the log-ratio induced by a convex mixture of the two. 


\input{figures/clip_sensitive}
\paragraph{Comparison to Skew RKL \citep{ko2024distillm}.} Although the RHS of Eq.~(\ref{eq:mixture}) is identical to skew RKL \citep{ko2024distillm}, our approach differs in application by using the bound $\frac{\log \lambda}{1-\lambda}$ strictly as a clipping threshold rather than modifying the global objective. \autoref{fig:clip_sensitivity} demonstrates that while Skew RKL is highly sensitive to $\lambda$ (e.g., dropping significantly at $\lambda=0.7$), our method remains robust. Remarkably, our lowest accuracy surpasses even the peak performance of Skew RKL. This confirms that selectively clipping heavy-tailed outliers stabilizes training more effectively than altering the global divergence.

%% file: figures/clip_sensitive.tex
\begin{wrapfigure}{r}{0.3\textwidth}
\vspace{0pt}
\centering
    \includegraphics[width=1.0\linewidth]{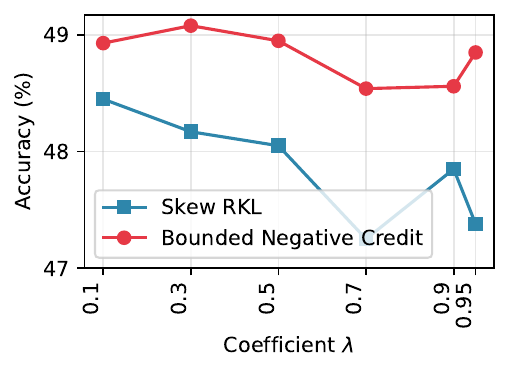}
    \vspace{-20pt}
    \caption{
        \textbf{Sensitivity analysis of $\lambda$.}
    }
    \vspace{-30pt}
    \label{fig:clip_sensitivity}
\end{wrapfigure}

%% file: sections/algorithm.tex
\clearpage
\section{Full Algorithm}
Here, we provide the algorithm for \alg in Algorithm~\ref{alg:main}.

\begin{algorithm}[h]
   \caption{\alg}\label{alg:main}
\begin{algorithmic}[1]
   \STATE {\bfseries Input:} student policy $\pi_{\theta}$, teacher policy $\pi_T$, query set $\mathcal{Q}$
   \STATE {\bfseries Hyperparameters:} total steps $K$, switch step $T_{\text{switch}}$, bounding coefficient $\lambda$, top-entropy fraction $\beta$, learning rate $\eta$
   \STATE {\bfseries Output:} trained student model $\pi_{\theta}$
   \STATE Initialize $\theta_{\text{old}} \leftarrow \theta$
   \FOR{$k = 1$ to $K$}
   \STATE Sample a batch of queries $\mathcal{B} \sim \mathcal{Q}$
   \STATE Generate $\{o_i\}_{i=1}^{G} \sim \pi_{\theta_{\text{old}}}(\cdot \mid q)$ for each $q \in \mathcal{B}$
   \STATE Compute $R_{i,t}(\theta) \leftarrow \log \frac{\pi_T(o_{i,t}|q, o_{i,<t})}{\pi_{\theta}(o_{i,t}|q, o_{i,<t})}$
   
   \STATE Clip \textcolor{blue}{$\hat{R}_{i,t}^{\lambda}(\theta) \leftarrow \max \left( \text{sg}(R_{i,t}(\theta)), \frac{\log \lambda}{1-\lambda}\right)$} \hfill $\triangleright$ Eq.~(\ref{eq:clip})
   
   \IF{$k < T_{\text{switch}}$}
       \STATE \textcolor{electriccrimson}{\textit{Phase I: Exploration (Credit-based Filtering)}}
       \STATE Set mask \textcolor{electriccrimson}{$M_{i,t}^{(k)} \leftarrow \mathbb{I}\left[ R_{i,t}(\theta) \ge \frac{\log \lambda}{1-\lambda} \right]$} \hfill $\triangleright$ Eq.~(\ref{eq:exploration_mask})
   \ELSE
       \STATE \textit{\textcolor{electriccrimson}{Phase II: Refinement} \textcolor{darkpastelgreen}{(Entropy-Guided Credit Allocation)}}
       \STATE Compute entropy $H_t^i$ for each token $o_{i,t}$ in batch
       \STATE Compute $\tau_\beta$ as the $(1-\beta)$-quantile of $\{H^i_t\}$
       \STATE Set mask \textcolor{darkpastelgreen}{$M_{i,t}^{(k)} \leftarrow \mathbb{I}\left[ H_{t}^i \ge \tau_\beta \right]$}  \hfill $\triangleright$ Eq.~(\ref{eq:refinement_mask})
   \ENDIF

   \STATE Compute gradients $\nabla_\theta \mathcal{J}_{\text{\alg}}$ using Eq.~(\ref{eq:unified_objective})
   \STATE Update parameters $\theta \leftarrow \theta + \eta \nabla_\theta \mathcal{J}_{\text{\alg}}$
   \STATE Update old policy parameters $\theta_{\text{old}} \leftarrow \theta$
   \ENDFOR
\end{algorithmic}
\end{algorithm}

%% file: sections/experimental-setup.tex
\clearpage
\section{Detailed Experimental Setup}\label{app:training}

\input{tables/hyperparams}
Section~\ref{sec:exp} details the experimental setup. We list the hyperparameter values for each setting in \autoref{tab:hyperparams}. Sensitivity analyses for the additive hyperparameters $\lambda$ and $\beta$ are presented in \autoref{fig:clip_sensitivity} and \autoref{fig:entropy_3b_7b}, respectively. Regarding the switch step $T_{\text{switch}}$ introduced in Section~\ref{sec:multi_stage}, we did not perform specific hyperparameter tuning; instead, we set the switch step to approximately one-quarter to one-third of the total training horizon, without extensive task-specific tuning. For $K$ and $\eta$, we adopted the same values as our baselines, where $\eta$ was determined via hyperparameter tuning based on vanilla RKL results.

\paragraph{Math reasoning.} We employ Verl~\citep{sheng2025hybridflow} for OPD. During rollout, we sample $n=8$ responses per prompt with a maximum response length of 8192 and a sampling temperature of 1.0. The global batch size is set to 128 with a mini-batch size of 32, resulting in 4 gradient updates per rollout step. The student policy is trained for 300 iterations. We utilize the AdamW optimizer~\citep{loshchilov2018decoupled} with a constant learning rate of $1\times10^{-5}$. All training runs are conducted on a single node equipped with 8$\times$ NVIDIA H100 80GB GPUs, requiring approximately 200 and 312 GPU hours for the 1.5B and 7B models, respectively. For evaluation, we use maximum response length of 32768 and a sampling temperature of 0.6 and top-p of 0.95. Our evaluation protocol follows the setup established by Qwen2.5-Math~\citep{yang2024qwen2} by using Verl~\citep{sheng2025hybridflow} for implementation. We employ training\footnote{\url{https://huggingface.co/datasets/Elliott/Openr1-Math-46k-8192}} and evaluation\footnote{\url{https://github.com/ElliottYan/LUFFY/blob/main/data/valid.parquet}} data from LUFFY~\citep{yan2025learning}.

\paragraph{Visual reasoning.} For visual tasks, we conduct OPD using Verl~\citep{sheng2025hybridflow}. We generate $n=12$ responses for each prompt, enforcing a maximum response length of 2048 and a sampling temperature of 1.0. Following the protocol in \citet{liu2025noisyrollout}, the models are trained for 60 iterations with a batch size of 128 and a mini-batch size of 64 (equating to 2 gradient updates per step). Optimization is performed via AdamW~\citep{loshchilov2018decoupled} with a learning rate of $5\times10^{-6}$. Using the same hardware setup (a single 8$\times$ NVIDIA H100 node), the training takes roughly 20 and 24 GPU hours for the 3B and 7B models. For evaluation, we use maximum response length of 8192 and a sampling temperature of 0.6 and top-p of 0.95 for nucleus sampling. We utilize the training\footnote{\url{https://huggingface.co/datasets/xyliu6/geometry3k}} and evaluation\footnote{\url{https://huggingface.co/datasets/xyliu6/noisyrollout_evaluation_data}} data from NoisyRollout~\citep{liu2025noisyrollout}.

\paragraph{Agentic reasoning with visual tool-use.} We implement OPD and RL based on the VerlTool framework~\citep{jiang2025verltool}. For rollout, the policy samples $n=8$ trajectories per prompt with a maximum response length of 8192 and a temperature of 1.0. We set the maximum number of tool-use rounds to 2. We maintain a batch size of 128 and a mini-batch size of 64, corresponding to 2 gradient updates per rollout step. The student policy undergoes training for 40 iterations using the AdamW optimizer~\citep{loshchilov2018decoupled} with a constant learning rate of $1\times10^{-6}$. The entire process consumes approximately 120 GPU hours on a single node with 8$\times$ NVIDIA H100 80GB GPUs. Following \citet{su2025pixel}, we set the maximum number of tool-use rounds to 5 during evaluation, with a maximum response length of 8192, a sampling temperature of 1.0, and a top-p of 1.0. For training, we utilize the PixelReasoner-RL dataset\footnote{\url{https://huggingface.co/datasets/TIGER-Lab/PixelReasoner-RL-Data}}. For evaluation, we employ the InfoVQA\footnote{\url{https://huggingface.co/datasets/JasperHaozhe/InfoVQA-EvalData-PixelReasoner}}, TallyQA\footnote{\url{https://huggingface.co/datasets/JasperHaozhe/TallyQA-EvalData-PixelReasoner}}, and VStar\footnote{\url{https://huggingface.co/datasets/JasperHaozhe/VStar-EvalData-PixelReasoner}} datasets provided by the PixelReasoner~\citep{su2025pixel}.

%% file: tables/hyperparams.tex
\begin{wraptable}{r}{0.6\textwidth}
\vspace{-12.5pt}
\centering
\caption{Hyperparameter values used in \alg experiments in Section~\ref{sec:exp}.}
\label{tab:hyperparams}
\resizebox{0.6\columnwidth}{!}{%
\begin{tabular}{l|ccccc}
\toprule
 & \multicolumn{2}{c}{\textbf{Math} (Section~\ref{sec:math})} & \multicolumn{2}{c}{\textbf{Visual} (Section~\ref{sec:visual})} & \textbf{Agentic} (Section~\ref{sec:tool}) \\
\midrule
Model Size & 1.5B & 7B & 3B & 7B & 3B \\
\midrule
Total step $K$ & 300 & 200 & 60 & 60 & 40 \\
Learning rate $\eta$ & $1\times10^{-5}$ & $3\times10^{-6}$ & $5\times10^{-6}$ & $5\times10^{-6}$ & $1\times10^{-6}$ \\
Clipping coefficient $\lambda$ & 0.3 & 0.3 & 0.3 & 0.3 & 0.3 \\
Top-entropy fraction $\beta$ & 0.2 & 0.2 & 0.2 & 0.2 & 0.4 \\
Switch step $T_{\text{switch}}$ & 100 & 50 & 20 & 20 & 10 \\
\bottomrule
\end{tabular}}%
\end{wraptable}

%% file: sections/extended_experiments.tex
\clearpage
\section{Additional Analyses and Discussions}\label{app:analysis}
In this section, we provide comprehensive analyses to offer a deeper understanding of the inner workings and robustness of \alg. Unless otherwise specified, all experiments follow the visual reasoning evaluation protocols introduced in Section~\ref{sec:visual}.

\input{tables/visual-diff}

\paragraph{Training with different teacher.} We further evaluate our approach by distilling from different teacher models to assess generalization. As detailed in \autoref{tab:visual-diff}, we utilize Qwen2.5-VL-7B-Instruct and 32B-Instruct models—fine-tuned on Geometry3K and MMK12 respectively via NoisyRollout \citep{liu2025noisyrollout}—as teachers. The results demonstrate the consistent effectiveness of \alg. Compared to the RKL baseline, our method yields uniform improvements across all six benchmarks for both teacher settings. Specifically, we observe an average accuracy gain of 1.34\% with the 7B teacher and 0.93\% with the 32B teacher. This confirms that \alg is robust to variations in teacher architecture and domain-specific expertise, reliably enhancing the student's visual reasoning and perception capabilities.

\input{figures/latency}
\paragraph{Scalability with longer training.} We further evaluate \alg under an extended training setting with 300 optimization steps and a combined training dataset consisting of Geometry3K and MMK12, using Qwen2.5-VL-3B-Instruct (\autoref{tab:longer-3b}) and Qwen2.5-VL-7B-Instruct (\autoref{tab:longer-7b}). As shown in \autoref{tab:longer-3b} and \autoref{tab:longer-7b}, extending the training horizon generally improves \alg's performance, while standard OPD exhibits less consistent gains, including degradation on some benchmarks. These results highlight the superior scalability of \alg under longer training. It outperforms the RKL baseline in both model sizes, achieving the highest average accuracy of 54.36\% with the 3B model and 58.83\% with the 7B model. These results indicate that our method is capable of continuously refining its policy given more compute and data, leading to robust improvements in both visual reasoning and perception tasks.

\paragraph{Extended test-time scaling results.} We provide a comprehensive evaluation of the test-time scaling capabilities of \alg on Geometry3K~\citep{lu2021inter} and MathVerse~\citep{zhang2024mathverse} benchmarks. \autoref{fig:latency-extension} demonstrates that the superior scaling of \alg is not limited to coverage metrics (Pass@$K$). Both the 3B and 7B models maintain a consistent lead in Maj@$K$, a metric that measures consensus robustness. This confirms that our method fundamentally increases the probability of correct reasoning chains, rather than merely generating over diverse ``lucky guess" to boost Pass@$K$.

\input{figures/timing}
\paragraph{Training time breakdown.} We analyze training wall-clock time to quantify computational overhead as shown in \autoref{fig:timing}. Contrary to concerns about the teacher's cost in on-policy distillation, our breakdown reveals it accounts for only a moderate fraction (8\%--22\%) of the total step. Crucially, this relative cost is inversely proportional to generation length. In long-context math tasks (8192 tokens), student generation dominates the runtime (77.3\%), rendering the teacher's impact marginal (8.2\%). Even in shorter visual reasoning tasks (2048 tokens), where the teacher's share rises to 21.8\%, student generation remains the largest component of runtime rather than teacher supervision.

\input{tables/full_vocab}
\paragraph{Extension to Top-K OPD.} We further compare \alg with GKD \citep{agarwal2024onpolicy} and DistiLLM \cite{ko2024distillm} which are full-vocabulary OPD. As shown in \autoref{tab:full-vocab}, both GKD and DistiLLM with full vocabulary incur out-of-memory (OOM) errors since they need to store the value for 150K vocab size for both student and the teacher models. To alleviate this cost, we apply the commonly used Top-K ($K=5$) approximation \cite{li2026rethinking}; however, this approximation is less effective and yields lower efficiency than sampled-token approaches. Nevertheless, under the same Top-5 setting, \alg still outperforms Top-5 approximated GKD and DistiLLM, showing that its gains are not limited to sampled-token training. Importantly, applying either full-vocabulary scoring or the Top-5 approximation to \alg does not bring meaningful improvement over sampled-token \alg, suggesting that the proposed credit allocation strategy is already sufficient without costly vocabulary-level approximation.

\begin{figure}[!htbp]
\centering
\begin{minipage}[t]{0.49\linewidth}
\centering
\includegraphics[width=\linewidth]{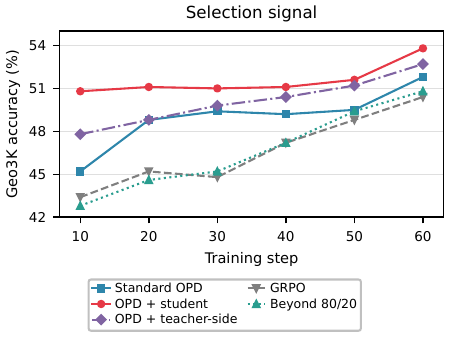}
\caption{\textbf{Effect of token-selection signals.} Comparison of student-entropy and teacher-entropy selection within OPD. Student-entropy selection achieves the highest accuracy throughout training.}
\label{fig:reb-entropy}
\end{minipage}\hfill
\begin{minipage}[t]{0.49\linewidth}
\centering
\includegraphics[width=\linewidth]{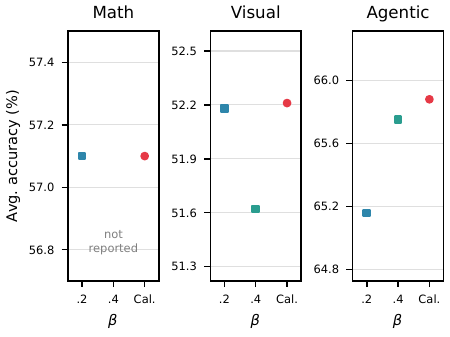}
\caption{\textbf{Sensitivity to the selection fraction.} Comparison of fixed and calibrated selection fractions across math, visual, and agentic reasoning. Entropy-based calibration achieves competitive performance without extensive tuning.}
\label{fig:reb-beta}
\end{minipage}
\end{figure}

\clearpage
\paragraph{Effect of the selection signal.} We compare different token-selection signals within OPD, along with GRPO and Beyond 80/20~\citep{wang2025beyond}. As shown in Figure~\ref{fig:reb-entropy}, student-entropy selection improves Geo3K accuracy from 45.2 to 50.8 at step 10 and from 51.8 to 53.8 at step 60, outperforming teacher-entropy selection and other baselines. Unlike EOPD~\citep{jin2026entropyaware}, which introduces an additional forward-KL term, the teacher-entropy variant modifies only the selection signal. These results highlight the effectiveness of student entropy for selective credit allocation.

\paragraph{Choice of the selection fraction.} We further examine the choice of $\beta$ by comparing fixed fractions with a calibration heuristic based on two-means clustering of token entropies from 16 sampled sequences. As shown in Figure~\ref{fig:reb-beta}, the calibrated fractions achieve comparable or slightly better performance than the best fixed choices across math, visual, and agentic reasoning. These results suggest that entropy-based calibration provides a practical heuristic for selecting $\beta$ without extensive tuning.

\begin{figure}[!htbp]
\centering
\includegraphics[width=\linewidth]{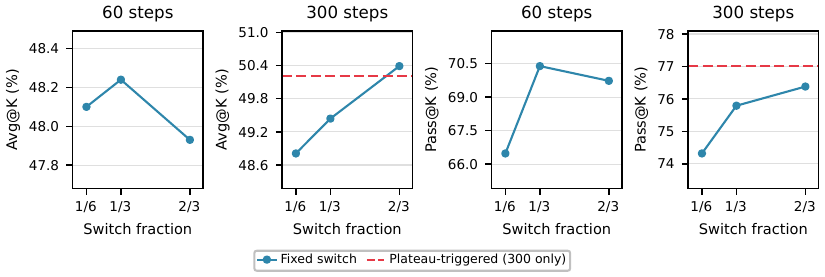}
\caption{\textbf{Sensitivity to switch timing.} We compare fixed switch fractions at 60 and 300 training steps on Geo3K and MathVerse. Switching at one-third performs best at 60 steps, while two-thirds performs best among fixed schedules at 300 steps. Red dashed lines indicate the plateau-triggered results at 300 steps.}
\label{fig:reb-switch}
\end{figure}

\paragraph{Effect of switch timing.} We analyze the sensitivity to switch timing by varying the switch fraction over ${1/6, 1/3, 2/3}$ for 60- and 300-step training. As shown in Figure~\ref{fig:reb-switch}, switching at one-third achieves the best performance at 60 steps, whereas two-thirds performs best among fixed schedules at 300 steps. Notably, the plateau-triggered schedule further improves Pass@$K$ at 300 steps, while the two-thirds switch achieves higher Avg@$K$. These results suggest that the preferred switch timing depends on the training horizon and evaluation metric.

\begin{figure}[!htbp]
\centering
\includegraphics[width=\linewidth]{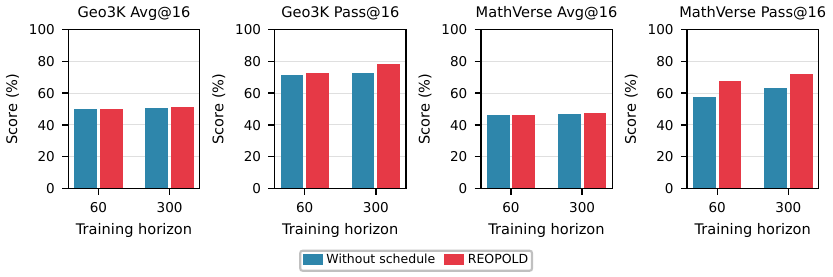}
\caption{\textbf{Effect of exploration-to-refinement scheduling.} We compare \alg with a variant that omits the initial exploration phase at 60 and 300 training steps. Scheduling consistently improves Pass@16 while yielding smaller gains in Avg@16, highlighting its contribution to solution coverage.}
\label{fig:reb-schedule}
\end{figure}

\paragraph{Effect on response coverage.}
We further examine the effectiveness of exploration-to-refinement scheduling by comparing \alg with a variant that removes the initial exploration phase while retaining bounded credit and entropy-guided token selection. As shown in Figure~\ref{fig:reb-schedule}, scheduling improves MathVerse Pass@16 by 10.54 and 9.08 points at 60 and 300 steps, respectively, while Avg@16 improves by only 0.46 and 1.16 points. These results suggest that the exploration phase primarily improves solution coverage, which is not fully captured by average accuracy.

%% file: tables/visual-diff.tex
\begin{table}[t]
\centering
\small
\begin{minipage}[t]{0.33\columnwidth}
\captionsetup{type=table}
\caption{Performance comparison of Qwen2.5-VL-3B-Instruct trained with various teacher models on visual reasoning reasoning and perception benchmarks. Accuracy (\%) is reported for all benchmarks.}
\label{tab:visual-diff}
\resizebox{\linewidth}{!}{
\begin{tabular}{lccc}
\toprule[0.1em]
           & RKL                   & \alg                       & $\Delta\%$                \\
\midrule
\rowcolor{blue!10} \multicolumn{4}{c}{\textbf{Qwen2.5-VL-7B-Instruct + NoisyRollout}}  \\
\midrule
\textbf{Geo3K}      & 49.75 & 51.41 & \textcolor{darkpastelgreen}{+1.66} \\
\textbf{MathVerse}  & 41.66 & 44.27 & \textcolor{darkpastelgreen}{+2.61} \\
\textbf{MathVision} & 23.33 & 24.01 & \textcolor{darkpastelgreen}{+0.68} \\
\textbf{MathVista}  & 62.20 & 63.10 & \textcolor{darkpastelgreen}{+0.90} \\
\textbf{WeMath}     & 64.83 & 65.66 & \textcolor{darkpastelgreen}{+0.83} \\
\textbf{Hallusion}  & 61.30 & 62.67 & \textcolor{darkpastelgreen}{+1.37} \\
\midrule
\textbf{AVG.}       & 50.51 & 51.85 & \textcolor{darkpastelgreen}{+1.34} \\
\midrule
\rowcolor{blue!10} \multicolumn{4}{c}{\textbf{Qwen2.5-VL-32B-Instruct + NoisyRollout$^\ddagger$}} \\
\midrule
\textbf{Geo3K}      & 43.93 & 45.76 & \textcolor{darkpastelgreen}{+1.83} \\
\textbf{MathVerse}  & 41.99 & 43.22 & \textcolor{darkpastelgreen}{+1.23} \\
\textbf{MathVision} & 25.13 & 25.79 & \textcolor{darkpastelgreen}{+0.66} \\
\textbf{MathVista}  & 63.60 & 64.20 & \textcolor{darkpastelgreen}{+0.60} \\
\textbf{WeMath}     & 64.02 & 64.66 & \textcolor{darkpastelgreen}{+0.64} \\
\textbf{Hallusion}  & 63.72 & 64.35 & \textcolor{darkpastelgreen}{+0.63} \\
\midrule
\textbf{AVG.}       & 50.40 & 51.33 & \textcolor{darkpastelgreen}{+0.93} \\
\bottomrule[0.1em]
\end{tabular}}
\end{minipage}
\hfill
\begin{minipage}[t]{0.32\columnwidth}
\captionsetup{type=table}
\caption{Performance comparison of Qwen2.5-VL-3B-Instruct trained for 60 and 300 training steps on visual reasoning reasoning and perception benchmarks. Accuracy (\%) is reported for all benchmarks.}
\label{tab:longer-3b}
\resizebox{\linewidth}{!}{
\begin{tabular}{lccc}
\toprule[0.1em]
           & RKL                   & \alg                       & $\Delta\%$                \\
\midrule
\rowcolor{blue!10} \multicolumn{4}{c}{\textbf{60 Training Steps (3B)}}  \\
\midrule
\textbf{Geo3K}      & 48.09 & 50.58 & \textcolor{darkpastelgreen}{+2.49} \\
\textbf{MathVerse}  & 33.25 & 46.40 & \textcolor{darkpastelgreen}{+13.15} \\
\textbf{MathVision} & 24.78 & 26.39 & \textcolor{darkpastelgreen}{+1.61} \\
\textbf{MathVista}  & 62.50 & 61.50 & \textcolor{electriccrimson}{-1.00} \\
\textbf{WeMath}     & 64.48 & 64.60 & \textcolor{darkpastelgreen}{+0.12} \\
\textbf{Hallusion}  & 60.83  & 63.62 & \textcolor{darkpastelgreen}{+2.79} \\
\midrule
\textbf{AVG.}       & 48.99 & 52.18 & \textcolor{darkpastelgreen}{+3.19} \\
\midrule
\rowcolor{blue!10} \multicolumn{4}{c}{\textbf{300 Training Steps (3B)}} \\
\midrule
\textbf{Geo3K}      & 49.08 & 51.08 & \textcolor{darkpastelgreen}{+2.00} \\
\textbf{MathVerse}  & 46.60 & 47.79 & \textcolor{darkpastelgreen}{+1.19} \\
\textbf{MathVision} & 26.16 & 27.44 & \textcolor{darkpastelgreen}{+1.28} \\
\textbf{MathVista}  & 67.20 & 66.30 & \textcolor{electriccrimson}{-0.90} \\
\textbf{WeMath}     & 66.43 & 67.18 & \textcolor{darkpastelgreen}{+0.75} \\
\textbf{Hallusion}  & 65.19 & 66.35 & \textcolor{darkpastelgreen}{+1.16} \\
\midrule
\textbf{AVG.}       & 53.44 & 54.36 & \textcolor{darkpastelgreen}{+0.92} \\
\bottomrule[0.1em]
\end{tabular}}
\end{minipage}
\hfill
\begin{minipage}[t]{0.32\columnwidth}
\captionsetup{type=table}
\caption{Performance comparison of Qwen2.5-VL-7B-Instruct trained for 60 and 300 training steps on visual reasoning reasoning and perception benchmarks. Accuracy (\%) is reported for all benchmarks.}
\label{tab:longer-7b}
\resizebox{\linewidth}{!}{
\begin{tabular}{lccc}
\toprule[0.1em]
           & RKL                   & \alg                       & $\Delta\%$                \\
\midrule
\rowcolor{blue!10} \multicolumn{4}{c}{\textbf{60 Training Steps (7B)}}  \\
\midrule
\textbf{Geo3K}      & 51.75 & 53.58 & \textcolor{darkpastelgreen}{+1.83} \\
\textbf{MathVerse}  & 47.71 & 51.43 & \textcolor{darkpastelgreen}{+3.72} \\
\textbf{MathVision} & 28.79 & 29.21 & \textcolor{darkpastelgreen}{+0.42} \\
\textbf{MathVista}  & 71.27 & 72.40 & \textcolor{darkpastelgreen}{+1.13} \\
\textbf{WeMath}     & 70.06 & 69.77 & \textcolor{electriccrimson}{-0.29} \\
\textbf{Hallusion}  & 69.51 & 70.14 & \textcolor{darkpastelgreen}{+0.63} \\
\midrule
\textbf{AVG.}       & 56.51 & 57.76 & \textcolor{darkpastelgreen}{+1.25} \\
\midrule
\rowcolor{blue!10} \multicolumn{4}{c}{\textbf{300 Training Steps (7B)}} \\
\midrule
\textbf{Geo3K}      & 49.42 & 53.58 & \textcolor{darkpastelgreen}{+4.16} \\
\textbf{MathVerse}  & 50.72 & 51.97 & \textcolor{darkpastelgreen}{+1.25} \\
\textbf{MathVision} & 29.74 & 31.12 & \textcolor{darkpastelgreen}{+1.38} \\
\textbf{MathVista}  & 71.20 & 73.60 & \textcolor{darkpastelgreen}{+2.40} \\
\textbf{WeMath}     & 69.43 & 71.84 & \textcolor{darkpastelgreen}{+2.41} \\
\textbf{Hallusion}  & 69.72 & 70.87 & \textcolor{darkpastelgreen}{+1.15} \\
\midrule
\textbf{AVG.}       & 56.71 & 58.83 & \textcolor{darkpastelgreen}{+2.12} \\
\bottomrule[0.1em]
\end{tabular}}
\end{minipage}
\end{table}

%% file: figures/latency.tex
\begin{figure*}[t]
    \centering
    \includegraphics[width=1.0\linewidth]{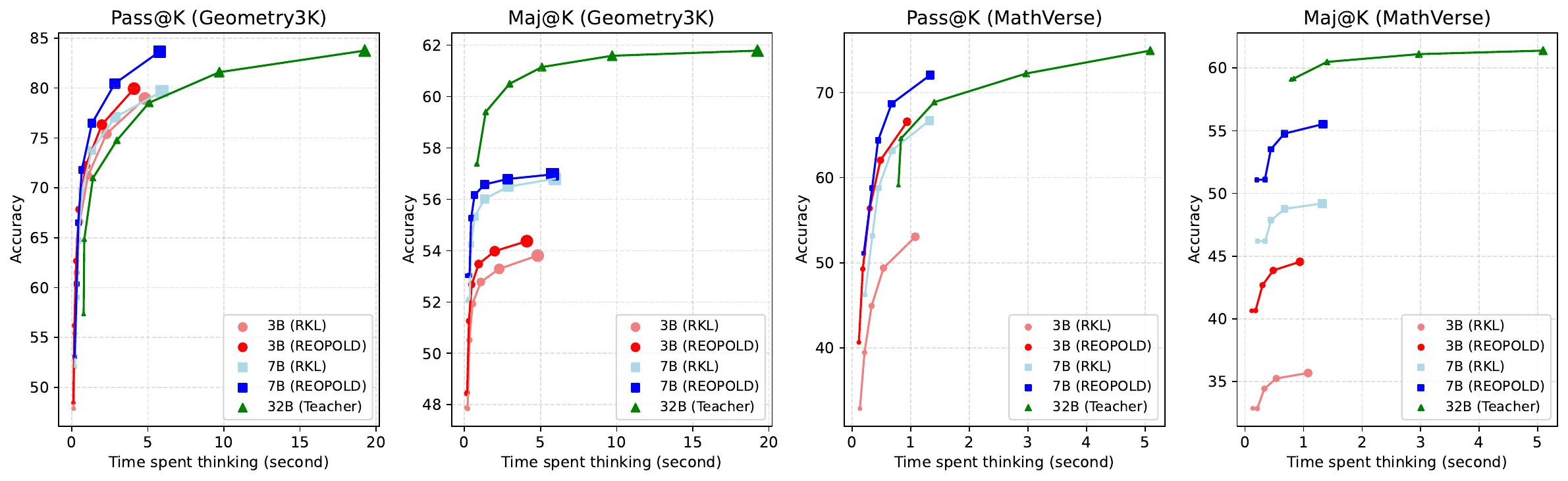}
    \caption{
        \textbf{Extended results of \autoref{fig:teaser}(b).} We visualize the accuracy (Pass@$K$ and Maj@$K$) against inference latency as the sample budget $K$ increases (up to 64 for Geometry3K; 16 for MathVerse). \alg (solid lines) consistently yields a better trade-off than the RKL baseline (faded lines). Notably, the 7B student matches or beats the 32B teacher's accuracy with significantly lower latency, confirming the efficiency of our distillation.
    }
    \vspace{-10pt}
    \label{fig:latency-extension}
\end{figure*}

%% file: figures/timing.tex
\begin{wrapfigure}{r}{0.45\textwidth}
\vspace{-15pt}
\centering
    \includegraphics[width=1.0\linewidth]{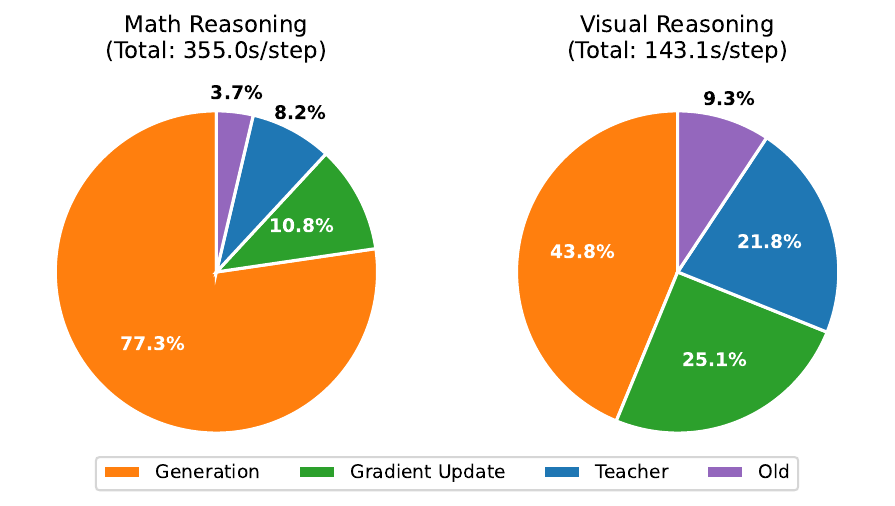}
    \caption{
        Breakdown of training wall-clock time per step.
    }
    \vspace{-10pt}
    \label{fig:timing}
\end{wrapfigure}

%% file: tables/full_vocab.tex
\begin{wraptable}{r}{0.55\textwidth}
\vspace{-12.5pt}
\centering
\caption{Comparison with full-vocabulary OPD and Top-K ($K=5$) approximations. Full-vocabulary GKD and DistiLLM (DLM) runs incur OOM under our visual reasoning setup; therefore, we additionally report Top-5 approximations.}
\label{tab:full-vocab}
\resizebox{1.0\linewidth}{!}{
\begin{tabular}{lccccc}
\toprule[0.1em]
           & GKD/DLM & GKD$^\dagger$ \cite{agarwal2024onpolicy} & DLM$^\dagger$ \cite{ko2024distillm} & \alg & \alg$^\dagger$ \\
\midrule
\textbf{Geo3K}      & OOM & 51.85 & 52.68 & \textbf{53.58} & 53.41 \\
\textbf{MathVerse}  & OOM & 48.14 & 48.65 & \textbf{51.97} & 50.96 \\
\textbf{MathVision} & OOM & 29.90 & 30.12 & \textbf{31.12} & 30.94 \\
\textbf{MathVista}  & OOM & 71.25 & 72.12 & 73.60 & \textbf{73.64} \\
\textbf{WeMath}     & OOM & 70.87 & 70.92 & \textbf{71.84} & 71.80 \\
\textbf{Hallusion}  & OOM & 70.21 & 70.37 & 70.87 & \textbf{71.52} \\
\midrule
\textbf{AVG.}       & OOM & 57.04 & 57.48 & \textbf{58.83} & 58.71 \\
\bottomrule[0.1em]
\end{tabular}}
\end{wraptable}

%% file: sections/qualitative.tex
\clearpage
\section{Qualitative Evaluation}

\begin{figure}[H]
    \centering
    \begin{tcolorbox}[colback=white, colframe=black!75, boxrule=0.8pt, arc=0pt, left=8pt, right=8pt, top=8pt, bottom=8pt]
        
        \ttfamily 
        \small 
        \raggedright
        
        \textbf{Question:} From 2001 to 2010, Is 2010 exported the most value of bread, biscuits, wafers, cakes and pastries from the UK? Your final answer can only be \boxed{yes} or \boxed{no}.
        
        \vspace{10pt}
        
        { 
            \centering
            \includegraphics[width=0.55\linewidth]{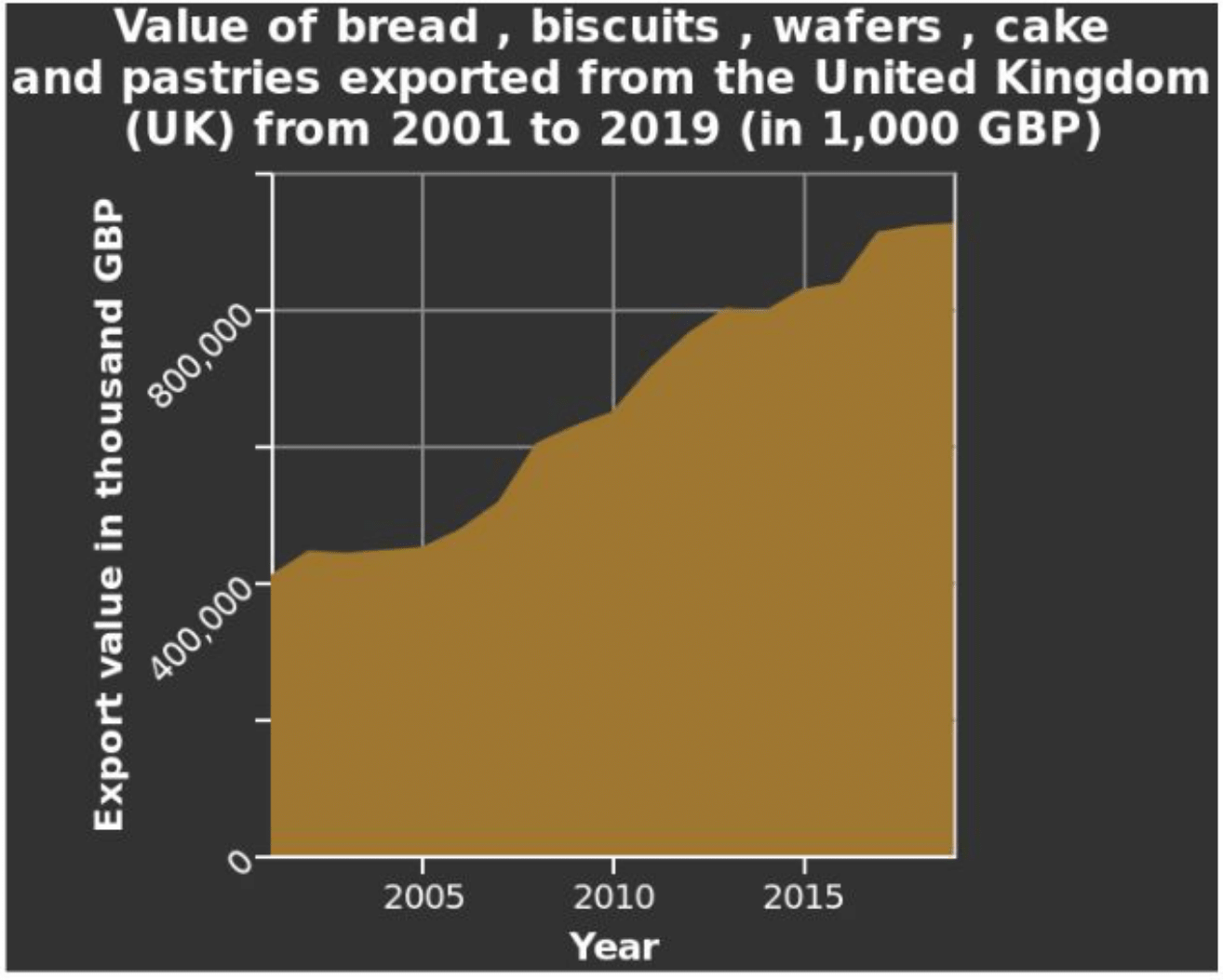} 
            \par 
        }
        
        \vspace{10pt}
        \noindent\rule{\textwidth}{0.4pt}
        \vspace{5pt}
        
        \textbf{[Qwen2.5-VL-3B-Instruct (\textnormal{\alg})]}
        \vspace{3pt}
        
        \textbf{\textless think\textgreater} \\
        To determine if 2010 exported the most value... we need to analyze the graph step by step. \\
        1. \textbf{Identify the Time Period}: The graph shows data from 2001 to 2019 ... \\
        2. \textbf{Analyze the Graph}: The graph is a line graph ... \\
        6. \textbf{Conclusion}: Based on the graph, \textcolor{darkpastelgreen}{\textbf{the export value in 2010 is the highest}} ... \\
        \textbf{\textless /think\textgreater} \\
        \vspace{3pt}
        Final Answer: \boxed{yes}
        
        \vspace{8pt}
        \noindent\rule{\textwidth}{0.4pt}
        \vspace{5pt}
        
        \textbf{[Qwen2.5-VL-3B-Instruct (Standard OPD)]}
        \vspace{3pt}
        
        \textbf{\#\#\# Step-by-Step Reasoning:} \\
        1. \textbf{Identify the Time Period}: The question asks about ... \\
        4. \textbf{Determine the Maximum Value}: ... \textcolor{electriccrimson}{\textbf{values are relatively close to 400,000 GBP for each year from 2001 to 2010. There is no clear indication that 2010 had a significantly higher}} ... \\
        5. \textbf{Conclusion}: Therefore, it is not possible to definitively say ... \\
        \vspace{3pt}
        \textbf{\#\#\# Final Answer:} \boxed{No}
    \end{tcolorbox}
    \caption{\textbf{Qualitative comparison on Hallusion Bench.} While the baseline trained with Standard OPD suffers from visual perception degradation (hallucinating a flat trend despite the visual evidence), \alg maintains robust visual grounding, accurately identifying the peak in the chart.}
    \label{fig:qualitative_code_style}
\end{figure}

\clearpage
\begin{figure}[H]
    \vspace{0pt}
    \centering
    \begin{tcolorbox}[colback=white, colframe=black!75, boxrule=0.8pt, arc=0pt, left=8pt, right=8pt, top=8pt, bottom=8pt]
        
        \ttfamily 
        \small 
        \raggedright
        
        \textbf{Question:} As shown in the figure, angle C = 36.0, angle BAE = 60.0, then the degree of angle AFC is () \\
        Choices: A:84$^{\circ}$ B:80$^{\circ}$ C:60$^{\circ}$ D:90$^{\circ}$
        
        
        { 
            \centering
            \includegraphics[width=0.2\linewidth]{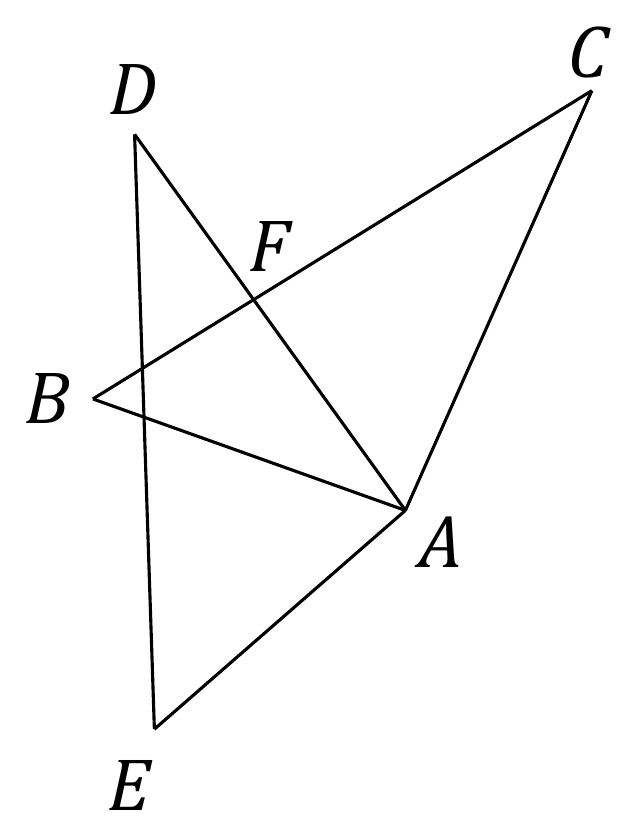} 
            \par 
        }
        
        \vspace{10pt}
        \noindent\rule{\textwidth}{0.4pt}
        \vspace{5pt}
        
        \textbf{[Qwen2.5-VL-7B-Instruct (\textnormal{\alg})]}
        \vspace{3pt}
        
        \textbf{\textless think\textgreater} \\
        To determine the measure of angle AFC, let's analyze the given information...
        
        4. \textbf{Analyze the angle AFC:} \\
        ... Substituting the given values: \\
        \[ \text{Angle AFC} = 60^{\circ} + 36^{\circ} = 96^{\circ} \]
        
        \textcolor{darkpastelgreen}{\textbf{However, we need to re-evaluate the problem}} because the angle AFC is actually the angle formed by the intersection...
        
        5. \textcolor{darkpastelgreen}{\textbf{Re-evaluate using the properties of the angles:}} \\
        ... The correct measure of angle AFC is: \\
        \[ \text{Angle AFC} = 180^{\circ} - 60^{\circ} - 36^{\circ} = 84^{\circ} \]
        \textbf{\textless /think\textgreater} \\
        \vspace{3pt}
        \boxed{A}
        
        \noindent\rule{\textwidth}{0.4pt}
        
        \textbf{[Qwen2.5-VL-7B-Instruct (Standard OPD)]}
        \vspace{3pt}
        
        \textbf{\textless think\textgreater} \\
        To determine the measure of angle AFC...
        
        5. \textbf{Consider the angle BAE:} \\
        ... Therefore: \\
        \[ 60^{\circ} = x + (144^{\circ} - x) \] \\
        This equation is not directly useful for finding $x$, but it confirms the relationship...
        
        ...
        
        8. \textbf{Final calculation:} \\
        The correct measure of angle AFC is: \\
        \[ \text{Angle AFC} = 156^{\circ} - 60^{\circ} \] \\
        \textbf{\[ \textcolor{electriccrimson}{\text{Angle AFC} = 90^{\circ}} \]}
        \textbf{\textless /think\textgreater} \\
        \vspace{3pt}
        \boxed{90}
        
    \end{tcolorbox}
    
    \caption{\textbf{Qualitative comparison on MathVerse.} The baseline (Standard OPD) gets trapped in circular logic (Step 5) and hallucinates the final calculation (Step 8). Our model initially derives an incorrect value ($96^{\circ}$) but explicitly triggers a self-correction process ("However, we need to re-evaluate...") to reach the correct solution ($84^{\circ}$).}
    \label{fig:qualitative_geometry_corrected}
    \label{fig:qualitative_geometry}
\end{figure}

\clearpage
\section{Limitations \& Impact Statement}
\paragraph{Limitations.} A limitation of \alg is that its gains may be smaller when the teacher and student are already closely matched, since most token-level credits become near zero and standard OPD is less unstable in this regime. In addition, \alg improves the allocation and stability of teacher feedback but does not directly reduce the cost of teacher inference; when teacher evaluation dominates wall-clock training time, the practical speedup may be limited. Future work could adapt the allocation rule to the teacher-student gap and combine selective credit allocation with cheaper teacher scoring or caching strategies.

\paragraph{Broader Impact.} This work improves the efficiency and stability of reasoning distillation for compact language models. By making sampled-token OPD more effective across mathematical, visual, and agentic tool-use tasks, \alg may reduce the data and optimization cost required to train capable smaller models, lowering deployment cost and latency in resource-constrained settings. However, stronger compact models may also lower the barrier to misuse, particularly in automated or tool-use applications. These risks are not unique to \alg and should be mitigated through safety evaluation, access control, and deployment monitoring.

\section{Reproducibility Statement}
For our experiments, we have clearly described the datasets used, the models fine-tuned, and the hyperparameters used (Section~\ref{sec:math}, Section~\ref{sec:visual}, Section~\ref{sec:tool}, and \autoref{app:training}). While all datasets and base models are publicly available, we provide a summary in \autoref{app:training}, along with relevant links. To further aid reproducibility, we provide a standalone implementation of our core algorithm in the supplementary material.

%% file: main.bib
@article{cui2025entropy,
  title   = {The entropy mechanism of reinforcement learning for reasoning language models},
  author  = {Cui, Ganqu and Zhang, Yuchen and Chen, Jiacheng and Yuan, Lifan and Wang, Zhi and Zuo, Yuxin and Li, Haozhan and Fan, Yuchen and Chen, Huayu and Chen, Weize and others},
  journal = {arXiv preprint arXiv:2505.22617},
  year    = {2025}
}

@article{williams1992simple,
  title     = {Simple statistical gradient-following algorithms for connectionist reinforcement learning},
  author    = {Williams, Ronald J},
  journal   = {Machine learning},
  volume    = {8},
  number    = {3},
  pages     = {229--256},
  year      = {1992},
  publisher = {Springer}
}

@article{chen2025minimax,
  title   = {MiniMax-M1: Scaling Test-Time Compute Efficiently with Lightning Attention},
  author  = {Chen, Aili and Li, Aonian and Gong, Bangwei and Jiang, Binyang and Fei, Bo and Yang, Bo and Shan, Boji and Yu, Changqing and Wang, Chao and Zhu, Cheng and others},
  journal = {arXiv preprint arXiv:2506.13585},
  year    = {2025}
}

@article{zheng2025group,
  title   = {Group Sequence Policy Optimization},
  author  = {Zheng, Chujie and Liu, Shixuan and Li, Mingze and Chen, Xiong-Hui and Yu, Bowen and Gao, Chang and Dang, Kai and Liu, Yuqiong and Men, Rui and Yang, An and others},
  journal = {arXiv preprint arXiv:2507.18071},
  year    = {2025}
}

@article{liu2025prorl,
  title   = {Prorl: Prolonged reinforcement learning expands reasoning boundaries in large language models},
  author  = {Liu, Mingjie and Diao, Shizhe and Lu, Ximing and Hu, Jian and Dong, Xin and Choi, Yejin and Kautz, Jan and Dong, Yi},
  journal = {arXiv preprint arXiv:2505.24864},
  year    = {2025}
}

@article{yan2025learning,
  title   = {Learning to reason under off-policy guidance},
  author  = {Yan, Jianhao and Li, Yafu and Hu, Zican and Wang, Zhi and Cui, Ganqu and Qu, Xiaoye and Cheng, Yu and Zhang, Yue},
  journal = {arXiv preprint arXiv:2504.14945},
  year    = {2025}
}

@article{wang2025beyond,
  title   = {Beyond the 80/20 rule: High-entropy minority tokens drive effective reinforcement learning for llm reasoning},
  author  = {Wang, Shenzhi and Yu, Le and Gao, Chang and Zheng, Chujie and Liu, Shixuan and Lu, Rui and Dang, Kai and Chen, Xionghui and Yang, Jianxin and Zhang, Zhenru and others},
  journal = {arXiv preprint arXiv:2506.01939},
  year    = {2025}
}

@article{he2025deepmath,
  title   = {Deepmath-103k: A large-scale, challenging, decontaminated, and verifiable mathematical dataset for advancing reasoning},
  author  = {He, Zhiwei and Liang, Tian and Xu, Jiahao and Liu, Qiuzhi and Chen, Xingyu and Wang, Yue and Song, Linfeng and Yu, Dian and Liang, Zhenwen and Wang, Wenxuan and others},
  journal = {arXiv preprint arXiv:2504.11456},
  year    = {2025}
}

@article{he2025skywork,
  title   = {Skywork open reasoner 1 technical report},
  author  = {He, Jujie and Liu, Jiacai and Liu, Chris Yuhao and Yan, Rui and Wang, Chaojie and Cheng, Peng and Zhang, Xiaoyu and Zhang, Fuxiang and Xu, Jiacheng and Shen, Wei and others},
  journal = {arXiv preprint arXiv:2505.22312},
  year    = {2025}
}

@article{luo2025deepscaler,
  title   = {Deepscaler: Surpassing o1-preview with a 1.5 b model by scaling rl},
  author  = {Luo, Michael and Tan, Sijun and Wong, Justin and Shi, Xiaoxiang and Tang, William Y and Roongta, Manan and Cai, Colin and Luo, Jeffrey and Zhang, Tianjun and Li, Li Erran and others},
  journal = {Notion Blog},
  year    = {2025}
}

@article{su2025pixel,
  title   = {Pixel reasoner: Incentivizing pixel-space reasoning with curiosity-driven reinforcement learning},
  author  = {Su, Alex and Wang, Haozhe and Ren, Weiming and Lin, Fangzhen and Chen, Wenhu},
  journal = {arXiv preprint arXiv:2505.15966},
  year    = {2025}
}

@article{guo2025deepseek,
  title   = {Deepseek-r1: Incentivizing reasoning capability in llms via reinforcement learning},
  author  = {Guo, Daya and Yang, Dejian and Zhang, Haowei and Song, Junxiao and Zhang, Ruoyu and Xu, Runxin and Zhu, Qihao and Ma, Shirong and Wang, Peiyi and Bi, Xiao and others},
  journal = {arXiv preprint arXiv:2501.12948},
  year    = {2025}
}

@article{shao2024deepseekmath,
  title   = {Deepseekmath: Pushing the limits of mathematical reasoning in open language models},
  author  = {Shao, Zhihong and Wang, Peiyi and Zhu, Qihao and Xu, Runxin and Song, Junxiao and Bi, Xiao and Zhang, Haowei and Zhang, Mingchuan and Li, YK and Wu, Yang and others},
  journal = {arXiv preprint arXiv:2402.03300},
  year    = {2024}
}

@inproceedings{sheng2025hybridflow,
  title     = {Hybridflow: A flexible and efficient rlhf framework},
  author    = {Sheng, Guangming and Zhang, Chi and Ye, Zilingfeng and Wu, Xibin and Zhang, Wang and Zhang, Ru and Peng, Yanghua and Lin, Haibin and Wu, Chuan},
  booktitle = {Proceedings of the Twentieth European Conference on Computer Systems},
  pages     = {1279--1297},
  year      = {2025}
}

@article{qian2025toolrl,
  title   = {Toolrl: Reward is all tool learning needs},
  author  = {Qian, Cheng and Acikgoz, Emre Can and He, Qi and Wang, Hongru and Chen, Xiusi and Hakkani-T{\"u}r, Dilek and Tur, Gokhan and Ji, Heng},
  journal = {arXiv preprint arXiv:2504.13958},
  year    = {2025}
}

@article{wei2022chain,
  title   = {Chain-of-thought prompting elicits reasoning in large language models},
  author  = {Wei, Jason and Wang, Xuezhi and Schuurmans, Dale and Bosma, Maarten and Xia, Fei and Chi, Ed and Le, Quoc V and Zhou, Denny and others},
  journal = {Advances in neural information processing systems},
  volume  = {35},
  pages   = {24824--24837},
  year    = {2022}
}

@article{yao2023tree,
  title   = {Tree of thoughts: Deliberate problem solving with large language models},
  author  = {Yao, Shunyu and Yu, Dian and Zhao, Jeffrey and Shafran, Izhak and Griffiths, Tom and Cao, Yuan and Narasimhan, Karthik},
  journal = {Advances in neural information processing systems},
  volume  = {36},
  pages   = {11809--11822},
  year    = {2023}
}

@inproceedings{yao2023react,
  title     = {React: Synergizing reasoning and acting in language models},
  author    = {Yao, Shunyu and Zhao, Jeffrey and Yu, Dian and Du, Nan and Shafran, Izhak and Narasimhan, Karthik and Cao, Yuan},
  booktitle = {International Conference on Learning Representations (ICLR)},
  year      = {2023}
}

@article{jaech2024openai,
  title   = {Openai o1 system card},
  author  = {Jaech, Aaron and Kalai, Adam and Lerer, Adam and Richardson, Adam and El-Kishky, Ahmed and Low, Aiden and Helyar, Alec and Madry, Aleksander and Beutel, Alex and Carney, Alex and others},
  journal = {arXiv preprint arXiv:2412.16720},
  year    = {2024}
}

@article{hendrycks2020measuring,
  title   = {Measuring massive multitask language understanding},
  author  = {Hendrycks, Dan and Burns, Collin and Basart, Steven and Zou, Andy and Mazeika, Mantas and Song, Dawn and Steinhardt, Jacob},
  journal = {arXiv preprint arXiv:2009.03300},
  year    = {2020}
}

@article{chen2025towards,
  title   = {Towards reasoning era: A survey of long chain-of-thought for reasoning large language models},
  author  = {Chen, Qiguang and Qin, Libo and Liu, Jinhao and Peng, Dengyun and Guan, Jiannan and Wang, Peng and Hu, Mengkang and Zhou, Yuhang and Gao, Te and Che, Wanxiang},
  journal = {arXiv preprint arXiv:2503.09567},
  year    = {2025}
}

@article{yang2024qwen2,
  title   = {Qwen2. 5-math technical report: Toward mathematical expert model via self-improvement},
  author  = {Yang, An and Zhang, Beichen and Hui, Binyuan and Gao, Bofei and Yu, Bowen and Li, Chengpeng and Liu, Dayiheng and Tu, Jianhong and Zhou, Jingren and Lin, Junyang and others},
  journal = {arXiv preprint arXiv:2409.12122},
  year    = {2024}
}

@article{yang2025qwen3,
  title   = {Qwen3 Technical Report},
  author  = {Yang, An and Li, Anfeng and Yang, Baosong and Zhang, Beichen and Hui, Binyuan and Zheng, Bo and Yu, Bowen and Gao, Chang and Huang, Chengen and Lv, Chenxu and others},
  year    = {2025},
  journal = {arxiv preprint arXiv: 2505.09388}
}

@article{zhao2025geometric,
  title={Geometric-Mean Policy Optimization},
  author={Zhao, Yuzhong and Liu, Yue and Liu, Junpeng and Chen, Jingye and Wu, Xun and Hao, Yaru and Lv, Tengchao and Huang, Shaohan and Cui, Lei and Ye, Qixiang and others},
  journal={arXiv preprint arXiv:2507.20673},
  year={2025}
}

@inproceedings{
ko2025distillm,
title={Disti{LLM}-2: A Contrastive Approach Boosts the Distillation of {LLM}s},
author={Jongwoo Ko and Tianyi Chen and Sungnyun Kim and Tianyu Ding and Luming Liang and Ilya Zharkov and Se-Young Yun},
booktitle={Forty-second International Conference on Machine Learning},
year={2025},
url={https://openreview.net/forum?id=rc65N9xIrY}
}

@inproceedings{
ko2024distillm,
title={Disti{LLM}: Towards Streamlined Distillation for Large Language Models},
author={Jongwoo Ko and Sungnyun Kim and Tianyi Chen and Se-Young Yun},
booktitle={Forty-first International Conference on Machine Learning},
year={2024},
url={https://openreview.net/forum?id=lsHZNNoC7r}
}

@misc{lu2025onpolicy,
  author       = {Kevin Lu and others},
  title        = {On-Policy Distillation},
  year         = {2025},
  howpublished = {\url{https://thinkingmachines.ai/blog/on-policy-distillation/}},
  note         = {Thinking Machines Blog, accessed on 2025-10-27}
}

@misc{patiño2025unlocking,
  author       = {Carlos Miguel Patiño and Kashif Rasul and Quentin Gallouédec and Ben Burtenshaw and Sergio Paniego and Vaibhav Srivastav and Thibaud Frere and Ed Beeching and Lewis Tunstall and Leandro von Werra and Thomas Wolf},
  title        = {Unlocking On-Policy Distillation for Any Model Family},
  year         = {2025},
  howpublished = {\url{https://huggingface.co/spaces/HuggingFaceH4/on-policy-distillation/}},
  note         = {HuggingFace Blog, accessed on 2025-10-29}
}

@inproceedings{
agarwal2024onpolicy,
title={On-Policy Distillation of Language Models: Learning from Self-Generated Mistakes},
author={Rishabh Agarwal and Nino Vieillard and Yongchao Zhou and Piotr Stanczyk and Sabela Ramos Garea and Matthieu Geist and Olivier Bachem},
booktitle={The Twelfth International Conference on Learning Representations},
year={2024},
url={https://openreview.net/forum?id=3zKtaqxLhW}
}

@inproceedings{hsieh-etal-2023-distilling,
    title = "Distilling Step-by-Step! Outperforming Larger Language Models with Less Training Data and Smaller Model Sizes",
    author = "Hsieh, Cheng-Yu  and
      Li, Chun-Liang  and
      Yeh, Chih-kuan  and
      Nakhost, Hootan  and
      Fujii, Yasuhisa  and
      Ratner, Alex  and
      Krishna, Ranjay  and
      Lee, Chen-Yu  and
      Pfister, Tomas",
    editor = "Rogers, Anna  and
      Boyd-Graber, Jordan  and
      Okazaki, Naoaki",
    booktitle = "Findings of the Association for Computational Linguistics: ACL 2023",
    month = jul,
    year = "2023",
    address = "Toronto, Canada",
    publisher = "Association for Computational Linguistics",
    url = "https://aclanthology.org/2023.findings-acl.507/",
    doi = "10.18653/v1/2023.findings-acl.507",
    pages = "8003--8017"
}

@inproceedings{ho-etal-2023-large,
    title = "Large Language Models Are Reasoning Teachers",
    author = "Ho, Namgyu  and
      Schmid, Laura  and
      Yun, Se-Young",
    editor = "Rogers, Anna  and
      Boyd-Graber, Jordan  and
      Okazaki, Naoaki",
    booktitle = "Proceedings of the 61st Annual Meeting of the Association for Computational Linguistics (Volume 1: Long Papers)",
    month = jul,
    year = "2023",
    address = "Toronto, Canada",
    publisher = "Association for Computational Linguistics",
    url = "https://aclanthology.org/2023.acl-long.830/",
    doi = "10.18653/v1/2023.acl-long.830",
    pages = "14852--14882"
}

@article{bai2025qwen2,
  title={Qwen2. 5-vl technical report},
  author={Bai, Shuai and Chen, Keqin and Liu, Xuejing and Wang, Jialin and Ge, Wenbin and Song, Sibo and Dang, Kai and Wang, Peng and Wang, Shijie and Tang, Jun and others},
  journal={arXiv preprint arXiv:2502.13923},
  year={2025}
}

@inproceedings{
lewkowycz2022solving,
title={Solving Quantitative Reasoning Problems with Language Models},
author={Aitor Lewkowycz and Anders Johan Andreassen and David Dohan and Ethan Dyer and Henryk Michalewski and Vinay Venkatesh Ramasesh and Ambrose Slone and Cem Anil and Imanol Schlag and Theo Gutman-Solo and Yuhuai Wu and Behnam Neyshabur and Guy Gur-Ari and Vedant Misra},
booktitle={Advances in Neural Information Processing Systems},
editor={Alice H. Oh and Alekh Agarwal and Danielle Belgrave and Kyunghyun Cho},
year={2022},
url={https://openreview.net/forum?id=IFXTZERXdM7}
}

@inproceedings{he-etal-2024-olympiadbench,
    title = "{O}lympiad{B}ench: A Challenging Benchmark for Promoting {AGI} with Olympiad-Level Bilingual Multimodal Scientific Problems",
    author = "He, Chaoqun  and
      Luo, Renjie  and
      Bai, Yuzhuo  and
      Hu, Shengding  and
      Thai, Zhen  and
      Shen, Junhao  and
      Hu, Jinyi  and
      Han, Xu  and
      Huang, Yujie  and
      Zhang, Yuxiang  and
      Liu, Jie  and
      Qi, Lei  and
      Liu, Zhiyuan  and
      Sun, Maosong",
    editor = "Ku, Lun-Wei  and
      Martins, Andre  and
      Srikumar, Vivek",
    booktitle = "Proceedings of the 62nd Annual Meeting of the Association for Computational Linguistics (Volume 1: Long Papers)",
    month = aug,
    year = "2024",
    address = "Bangkok, Thailand",
    publisher = "Association for Computational Linguistics",
    url = "https://aclanthology.org/2024.acl-long.211/",
    doi = "10.18653/v1/2024.acl-long.211",
    pages = "3828--3850"
}

@inproceedings{wu-etal-2025-rethinking,
    title = "Rethinking {K}ullback-{L}eibler Divergence in Knowledge Distillation for Large Language Models",
    author = "Wu, Taiqiang  and
      Tao, Chaofan  and
      Wang, Jiahao  and
      Yang, Runming  and
      Zhao, Zhe  and
      Wong, Ngai",
    editor = "Rambow, Owen  and
      Wanner, Leo  and
      Apidianaki, Marianna  and
      Al-Khalifa, Hend  and
      Eugenio, Barbara Di  and
      Schockaert, Steven",
    booktitle = "Proceedings of the 31st International Conference on Computational Linguistics",
    month = jan,
    year = "2025",
    address = "Abu Dhabi, UAE",
    publisher = "Association for Computational Linguistics",
    url = "https://aclanthology.org/2025.coling-main.383/",
    pages = "5737--5755"
}

@inproceedings{
liu2025noisyrollout,
title={NoisyRollout: Reinforcing Visual Reasoning with Data Augmentation},
author={Xiangyan Liu and Jinjie Ni and Zijian Wu and Chao Du and Longxu Dou and Haonan Wang and Tianyu Pang and Michael Qizhe Shieh},
booktitle={The Thirty-ninth Annual Conference on Neural Information Processing Systems},
year={2025},
url={https://openreview.net/forum?id=9zD2i7YRot}
}

@inproceedings{zhang2024mathverse,
  title={Mathverse: Does your multi-modal llm truly see the diagrams in visual math problems?},
  author={Zhang, Renrui and Jiang, Dongzhi and Zhang, Yichi and Lin, Haokun and Guo, Ziyu and Qiu, Pengshuo and Zhou, Aojun and Lu, Pan and Chang, Kai-Wei and Qiao, Yu and others},
  booktitle={European Conference on Computer Vision},
  pages={169--186},
  year={2024},
  organization={Springer}
}

@article{wang2024measuring,
  title={Measuring multimodal mathematical reasoning with math-vision dataset},
  author={Wang, Ke and Pan, Junting and Shi, Weikang and Lu, Zimu and Ren, Houxing and Zhou, Aojun and Zhan, Mingjie and Li, Hongsheng},
  journal={Advances in Neural Information Processing Systems},
  volume={37},
  pages={95095--95169},
  year={2024}
}

@article{lu2023mathvista,
  title={Mathvista: Evaluating mathematical reasoning of foundation models in visual contexts},
  author={Lu, Pan and Bansal, Hritik and Xia, Tony and Liu, Jiacheng and Li, Chunyuan and Hajishirzi, Hannaneh and Cheng, Hao and Chang, Kai-Wei and Galley, Michel and Gao, Jianfeng},
  journal={arXiv preprint arXiv:2310.02255},
  year={2023}
}

@inproceedings{qiao2025we,
  title={We-math: Does your large multimodal model achieve human-like mathematical reasoning?},
  author={Qiao, Runqi and Tan, Qiuna and Dong, Guanting and MinhuiWu, MinhuiWu and Sun, Chong and Song, Xiaoshuai and Wang, Jiapeng and Gongque, Zhuoma and Lei, Shanglin and Zhang, Yifan and others},
  booktitle={Proceedings of the 63rd Annual Meeting of the Association for Computational Linguistics (Volume 1: Long Papers)},
  pages={20023--20070},
  year={2025}
}

@article{lu2021inter,
  title={Inter-gps: Interpretable geometry problem solving with formal language and symbolic reasoning},
  author={Lu, Pan and Gong, Ran and Jiang, Shibiao and Qiu, Liang and Huang, Siyuan and Liang, Xiaodan and Zhu, Song-Chun},
  journal={arXiv preprint arXiv:2105.04165},
  year={2021}
}

@article{wang2025perception,
  title={Perception-aware policy optimization for multimodal reasoning},
  author={Wang, Zhenhailong and Guo, Xuehang and Stoica, Sofia and Xu, Haiyang and Wang, Hongru and Ha, Hyeonjeong and Chen, Xiusi and Chen, Yangyi and Yan, Ming and Huang, Fei and others},
  journal={arXiv preprint arXiv:2507.06448},
  year={2025}
}

@article{team2023gemini,
  title={Gemini: a family of highly capable multimodal models},
  author={Team, Gemini and Anil, Rohan and Borgeaud, Sebastian and Alayrac, Jean-Baptiste and Yu, Jiahui and Soricut, Radu and Schalkwyk, Johan and Dai, Andrew M and Hauth, Anja and Millican, Katie and others},
  journal={arXiv preprint arXiv:2312.11805},
  year={2023}
}

@inproceedings{
Holtzman2020The,
title={The Curious Case of Neural Text Degeneration},
author={Ari Holtzman and Jan Buys and Li Du and Maxwell Forbes and Yejin Choi},
booktitle={International Conference on Learning Representations},
year={2020},
url={https://openreview.net/forum?id=rygGQyrFvH}
}

@inproceedings{guan2024hallusionbench,
  title={Hallusionbench: an advanced diagnostic suite for entangled language hallucination and visual illusion in large vision-language models},
  author={Guan, Tianrui and Liu, Fuxiao and Wu, Xiyang and Xian, Ruiqi and Li, Zongxia and Liu, Xiaoyu and Wang, Xijun and Chen, Lichang and Huang, Furong and Yacoob, Yaser and others},
  booktitle={Proceedings of the IEEE/CVF Conference on Computer Vision and Pattern Recognition},
  pages={14375--14385},
  year={2024}
}

@article{jiang2025verltool,
  title={Verltool: Towards holistic agentic reinforcement learning with tool use},
  author={Jiang, Dongfu and Lu, Yi and Li, Zhuofeng and Lyu, Zhiheng and Nie, Ping and Wang, Haozhe and Su, Alex and Chen, Hui and Zou, Kai and Du, Chao and others},
  journal={arXiv preprint arXiv:2509.01055},
  year={2025}
}

@inproceedings{acharya2019tallyqa,
  title={Tallyqa: Answering complex counting questions},
  author={Acharya, Manoj and Kafle, Kushal and Kanan, Christopher},
  booktitle={Proceedings of the AAAI conference on artificial intelligence},
  volume={33},
  pages={8076--8084},
  year={2019}
}

@inproceedings{wu2024v,
  title={V?: Guided visual search as a core mechanism in multimodal llms},
  author={Wu, Penghao and Xie, Saining},
  booktitle={Proceedings of the IEEE/CVF Conference on Computer Vision and Pattern Recognition},
  pages={13084--13094},
  year={2024}
}

@inproceedings{mathew2022infographicvqa,
  title={Infographicvqa},
  author={Mathew, Minesh and Bagal, Viraj and Tito, Rub{\`e}n and Karatzas, Dimosthenis and Valveny, Ernest and Jawahar, CV},
  booktitle={Proceedings of the IEEE/CVF Winter Conference on Applications of Computer Vision},
  pages={1697--1706},
  year={2022}
}

@inproceedings{
loshchilov2018decoupled,
title={Decoupled Weight Decay Regularization},
author={Ilya Loshchilov and Frank Hutter},
booktitle={International Conference on Learning Representations},
year={2019},
url={https://openreview.net/forum?id=Bkg6RiCqY7},
}

@inproceedings{
gu2024minillm,
title={Mini{LLM}: Knowledge Distillation of Large Language Models},
author={Yuxian Gu and Li Dong and Furu Wei and Minlie Huang},
booktitle={The Twelfth International Conference on Learning Representations},
year={2024},
url={https://openreview.net/forum?id=5h0qf7IBZZ}
}

@article{dang2025reinforcement,
  title={Reinforcement Learning for Reasoning in Small LLMs: What Works and What Doesn't},
  author={Dang, Quy-Anh and Ngo, Chris},
  journal={arXiv preprint arXiv:2503.16219},
  year={2025}
}

@article{xiao2026mimo,
  title={MiMo-V2-Flash Technical Report},
  author={Xiao, Bangjun and Xia, Bingquan and Yang, Bo and Gao, Bofei and Shen, Bowen and Zhang, Chen and He, Chenhong and Lou, Chiheng and Luo, Fuli and Wang, Gang and others},
  journal={arXiv preprint arXiv:2601.02780},
  year={2026}
}

@article{gudibande2023false,
  title={The false promise of imitating proprietary llms},
  author={Gudibande, Arnav and Wallace, Eric and Snell, Charlie and Geng, Xinyang and Liu, Hao and Abbeel, Pieter and Levine, Sergey and Song, Dawn},
  journal={arXiv preprint arXiv:2305.15717},
  year={2023}
}

@article{creswell2022selection,
  title={Selection-inference: Exploiting large language models for interpretable logical reasoning},
  author={Creswell, Antonia and Shanahan, Murray and Higgins, Irina},
  journal={arXiv preprint arXiv:2205.09712},
  year={2022}
}

@inproceedings{
li2024neuralsymbolic,
title={Neural-Symbolic Recursive Machine for Systematic Generalization},
author={Qing Li and Yixin Zhu and Yitao Liang and Ying Nian Wu and Song-Chun Zhu and Siyuan Huang},
booktitle={The Twelfth International Conference on Learning Representations},
year={2024},
url={https://openreview.net/forum?id=FWJAmwE0xH}
}

@inproceedings{huang-chang-2023-towards,
    title = "Towards Reasoning in Large Language Models: A Survey",
    author = "Huang, Jie  and
      Chang, Kevin Chen-Chuan",
    editor = "Rogers, Anna  and
      Boyd-Graber, Jordan  and
      Okazaki, Naoaki",
    booktitle = "Findings of the Association for Computational Linguistics: ACL 2023",
    month = jul,
    year = "2023",
    address = "Toronto, Canada",
    publisher = "Association for Computational Linguistics",
    url = "https://aclanthology.org/2023.findings-acl.67/",
    doi = "10.18653/v1/2023.findings-acl.67",
    pages = "1049--1065"
}

@inproceedings{
zhang2024chain,
title={Chain of Preference Optimization: Improving Chain-of-Thought Reasoning in {LLM}s},
author={Xuan Zhang and Chao Du and Tianyu Pang and Qian Liu and Wei Gao and Min Lin},
booktitle={The Thirty-eighth Annual Conference on Neural Information Processing Systems},
year={2024},
url={https://openreview.net/forum?id=2cczgOfMP4}
}

@inproceedings{
wang2023selfconsistency,
title={Self-Consistency Improves Chain of Thought Reasoning in Language Models},
author={Xuezhi Wang and Jason Wei and Dale Schuurmans and Quoc V Le and Ed H. Chi and Sharan Narang and Aakanksha Chowdhery and Denny Zhou},
booktitle={The Eleventh International Conference on Learning Representations },
year={2023},
url={https://openreview.net/forum?id=1PL1NIMMrw}
}

@inproceedings{
zhao2023large,
title={Large Language Models as Commonsense Knowledge for Large-Scale Task Planning},
author={Zirui Zhao and Wee Sun Lee and David Hsu},
booktitle={Thirty-seventh Conference on Neural Information Processing Systems},
year={2023},
url={https://openreview.net/forum?id=Wjp1AYB8lH}
}

@inproceedings{
calanzone2025logically,
title={Logically Consistent Language Models via Neuro-Symbolic Integration},
author={Diego Calanzone and Stefano Teso and Antonio Vergari},
booktitle={The Thirteenth International Conference on Learning Representations},
year={2025},
url={https://openreview.net/forum?id=7PGluppo4k}
}

@inproceedings{
weng2024mastering,
title={Mastering Symbolic Operations: Augmenting Language Models with Compiled Neural Networks},
author={Yixuan Weng and Minjun Zhu and Fei Xia and Bin Li and Shizhu He and Kang Liu and Jun Zhao},
booktitle={The Twelfth International Conference on Learning Representations},
year={2024},
url={https://openreview.net/forum?id=9nsNyN0vox}
}

@inproceedings{
barbiero2023interpretable,
title={Interpretable Neural-Symbolic Concept Reasoning},
author={Pietro Barbiero and Gabriele Ciravegna and Francesco Giannini and Mateo Espinosa Zarlenga and Lucie Charlotte Magister and Alberto Tonda and Pietro Lio and Frederic Precioso and Mateja Jamnik and Giuseppe Marra},
booktitle={ICML 2023 Workshop on Differentiable Almost Everything: Differentiable Relaxations, Algorithms, Operators, and Simulators},
year={2023},
url={https://openreview.net/forum?id=oRj82I2apn}
}

@inproceedings{hao-etal-2023-reasoning,
    title = "Reasoning with Language Model is Planning with World Model",
    author = "Hao, Shibo  and
      Gu, Yi  and
      Ma, Haodi  and
      Hong, Joshua  and
      Wang, Zhen  and
      Wang, Daisy  and
      Hu, Zhiting",
    editor = "Bouamor, Houda  and
      Pino, Juan  and
      Bali, Kalika",
    booktitle = "Proceedings of the 2023 Conference on Empirical Methods in Natural Language Processing",
    month = dec,
    year = "2023",
    address = "Singapore",
    publisher = "Association for Computational Linguistics",
    url = "https://aclanthology.org/2023.emnlp-main.507/",
    doi = "10.18653/v1/2023.emnlp-main.507",
    pages = "8154--8173",
}

@inproceedings{giadikiaroglou-etal-2024-puzzle,
    title = "Puzzle Solving using Reasoning of Large Language Models: A Survey",
    author = "Giadikiaroglou, Panagiotis  and
      Lymperaiou, Maria  and
      Filandrianos, Giorgos  and
      Stamou, Giorgos",
    editor = "Al-Onaizan, Yaser  and
      Bansal, Mohit  and
      Chen, Yun-Nung",
    booktitle = "Proceedings of the 2024 Conference on Empirical Methods in Natural Language Processing",
    month = nov,
    year = "2024",
    address = "Miami, Florida, USA",
    publisher = "Association for Computational Linguistics",
    url = "https://aclanthology.org/2024.emnlp-main.646/",
    doi = "10.18653/v1/2024.emnlp-main.646",
    pages = "11574--11591",
}

@article{santana2025question,
  title={Question Answering with LLMs and Learning from Answer Sets},
  author={Santana, Manuel Alejandro Borroto and Gallagher, Katie and Ielo, Antonio and Kareem, Irfan and Ricca, Francesco and Russo, Alessandra},
  journal={Theory and Practice of Logic Programming},
  pages={1--25},
  year={2025},
  publisher={Cambridge University Press}
}

@inproceedings{rajani-etal-2019-explain,
    title = "Explain Yourself! Leveraging Language Models for Commonsense Reasoning",
    author = "Rajani, Nazneen Fatema  and
      McCann, Bryan  and
      Xiong, Caiming  and
      Socher, Richard",
    editor = "Korhonen, Anna  and
      Traum, David  and
      M{\`a}rquez, Llu{\'i}s",
    booktitle = "Proceedings of the 57th Annual Meeting of the Association for Computational Linguistics",
    month = jul,
    year = "2019",
    address = "Florence, Italy",
    publisher = "Association for Computational Linguistics",
    url = "https://aclanthology.org/P19-1487/",
    doi = "10.18653/v1/P19-1487",
    pages = "4932--4942",
}

@article{xu2025kdrl,
  title={KDRL: Post-Training Reasoning LLMs via Unified Knowledge Distillation and Reinforcement Learning},
  author={Xu, Hongling and Zhu, Qi and Deng, Heyuan and Li, Jinpeng and Hou, Lu and Wang, Yasheng and Shang, Lifeng and Xu, Ruifeng and Mi, Fei},
  journal={arXiv preprint arXiv:2506.02208},
  year={2025}
}

@inproceedings{
ko2025flexjudge,
title={Flex-Judge: Text-Only Reasoning Unleashes Zero-Shot Multimodal Evaluators},
author={Jongwoo Ko and Sungnyun Kim and Sungwoo Cho and Se-Young Yun},
booktitle={The Thirty-ninth Annual Conference on Neural Information Processing Systems},
year={2025},
url={https://openreview.net/forum?id=v6kyF3S7dM}
}

@article{yang2026learning,
  title={Learning beyond teacher: Generalized on-policy distillation with reward extrapolation},
  author={Yang, Wenkai and Liu, Weijie and Xie, Ruobing and Yang, Kai and Yang, Saiyong and Lin, Yankai},
  journal={arXiv preprint arXiv:2602.12125},
  year={2026}
}

@article{xu2025phi,
  title={Phi-4-mini-reasoning: Exploring the limits of small reasoning language models in math},
  author={Xu, Haoran and Peng, Baolin and Awadalla, Hany and Chen, Dongdong and Chen, Yen-Chun and Gao, Mei and Kim, Young Jin and Li, Yunsheng and Ren, Liliang and Shen, Yelong and others},
  journal={arXiv preprint arXiv:2504.21233},
  year={2025}
}

@article{schaul2015prioritized,
  title={Prioritized experience replay},
  author={Schaul, Tom and Quan, John and Antonoglou, Ioannis and Silver, David},
  journal={arXiv preprint arXiv:1511.05952},
  year={2015}
}

@inproceedings{lin2017focal,
  title={Focal loss for dense object detection},
  author={Lin, Tsung-Yi and Goyal, Priya and Girshick, Ross and He, Kaiming and Doll{\'a}r, Piotr},
  booktitle={Proceedings of the IEEE international conference on computer vision},
  pages={2980--2988},
  year={2017}
}

@inproceedings{ren2018learning,
  title={Learning to reweight examples for robust deep learning},
  author={Ren, Mengye and Zeng, Wenyuan and Yang, Bin and Urtasun, Raquel},
  booktitle={International conference on machine learning},
  pages={4334--4343},
  year={2018},
  organization={PMLR}
}

@inproceedings{shrivastava2016training,
  title={Training region-based object detectors with online hard example mining},
  author={Shrivastava, Abhinav and Gupta, Abhinav and Girshick, Ross},
  booktitle={Proceedings of the IEEE conference on computer vision and pattern recognition},
  pages={761--769},
  year={2016}
}

@article{kakade2001natural,
  title={A natural policy gradient},
  author={Kakade, Sham M},
  journal={Advances in neural information processing systems},
  volume={14},
  year={2001}
}

@article{
boizard2025towards,
title={Towards Cross-Tokenizer Distillation: the Universal Logit Distillation Loss for {LLM}s},
author={Nicolas Boizard and Kevin El Haddad and CELINE HUDELOT and Pierre Colombo},
journal={Transactions on Machine Learning Research},
issn={2835-8856},
year={2025},
url={https://openreview.net/forum?id=bwRxXiGO9A},
note={}
}

@article{zhao2026self,
  title={Self-Distilled Reasoner: On-Policy Self-Distillation for Large Language Models},
  author={Zhao, Siyan and Xie, Zhihui and Liu, Mengchen and Huang, Jing and Pang, Guan and Chen, Feiyu and Grover, Aditya},
  journal={arXiv preprint arXiv:2601.18734},
  year={2026}
}

@article{shenfeld2026self,
  title={Self-Distillation Enables Continual Learning},
  author={Shenfeld, Idan and Damani, Mehul and H{\"u}botter, Jonas and Agrawal, Pulkit},
  journal={arXiv preprint arXiv:2601.19897},
  year={2026}
}

@article{li2026rethinking,
  title={Rethinking On-Policy Distillation of Large Language Models: Phenomenology, Mechanism, and Recipe},
  author={Li, Yaxuan and Zuo, Yuxin and He, Bingxiang and Zhang, Jinqian and Xiao, Chaojun and Qian, Cheng and Yu, Tianyu and Gao, Huan-ang and Yang, Wenkai and Liu, Zhiyuan and others},
  journal={arXiv preprint arXiv:2604.13016},
  year={2026}
}

@inproceedings{
fu2026revisiting,
title={Revisiting On-Policy Distillation: Empirical Failure Modes and Simple Fixes},
author={Yuqian Fu and Haohuan Huang and Kaiwen Jiang and Jiacai Liu and Zhuo Jiang and Yuanheng Zhu and Dongbin Zhao},
booktitle={Third Conference on Language Modeling},
year={2026},
url={https://openreview.net/forum?id=uZbZb058bA}
}

@inproceedings{
jin2026entropyaware,
title={Entropy-Aware On-Policy Distillation of Language Models},
author={Woogyeol Jin and Taywon Min and Yongjin Yang and Dennis Wei and Yi Zhou and Swanand Ravindra Kadhe and Nathalie Baracaldo and Kimin Lee},
booktitle={Forty-third International Conference on Machine Learning},
year={2026},
url={https://openreview.net/forum?id=J5i09faOOf}
}
